\documentclass[letterpaper, 10 pt, conference]{ieeeconf}  

\IEEEoverridecommandlockouts                              

\usepackage{graphics} 
\usepackage{epsfig} 
\usepackage{times} 
\usepackage{amsmath} 
\usepackage{amssymb}  
\usepackage{booktabs}
\usepackage{caption}
\usepackage{graphicx}
\usepackage{float} 
\usepackage{hyperref}
\usepackage{subcaption}
\usepackage{color}

\hypersetup{
    colorlinks=true,
    linkcolor=blue,
    filecolor=magenta,      
    urlcolor=cyan,
    pdftitle={Overleaf Example},
    pdfpagemode=FullScreen,
    }

\title{\LARGE \bf
NGEL-SLAM: Neural Implicit Representation-based \\ Global Consistent Low-Latency SLAM System
}

\author{Yunxuan Mao$^{1}$, Xuan Yu$^{1}$, Zhuqing Zhang$^{1}$, Kai Wang$^{2}$, Yue Wang$^{1}$, Rong Xiong$^{1}$, and Yiyi Liao$^{1*}$
\thanks{This work was supported by the National Key R\&D Program of China under Grant 2021ZD0114500.}
\thanks{$^{1}$Yunxuan Mao, Xuan Yu, Zhuqing Zhang, Yue Wang, Rong Xiong, and Yiyi Liao are with Zhejiang University, Hangzhou, China. }%
\thanks{$^{2}$Kai Wang is with the Application Innovate Lab, Huawei Incorporated Company, Beijing, China.}
\thanks{*Corresponding author.}
}

\begin{document}

\maketitle
\thispagestyle{empty}
\pagestyle{empty}


\begin{abstract}
Neural implicit representations have emerged as a promising solution for providing dense geometry in Simultaneous Localization and Mapping (SLAM). However, existing methods in this direction fall short in terms of global consistency and low latency. This paper presents NGEL-SLAM to tackle the above challenges. To ensure global consistency, our system leverages a traditional feature-based tracking module that incorporates loop closure. Additionally, we maintain a global consistent map by representing the scene using multiple neural implicit fields, enabling quick adjustment to the loop closure. Moreover, our system allows for fast convergence through the use of octree-based implicit representations. The combination of rapid response to loop closure and fast convergence makes our system a truly low-latency system that achieves global consistency. Our system enables rendering high-fidelity RGB-D images, along with extracting dense and complete surfaces. Experiments on both synthetic and real-world datasets suggest that our system achieves state-of-the-art tracking and mapping accuracy while maintaining low latency. The code is available at \url{https://github.com/YunxuanMao/ngel_slam}.

\end{abstract}


\section{Introduction}
Dense visual simultaneous localization and mapping (SLAM) is a fundamental and challenging problem in computer vision. It involves updating a map of an unknown environment while simultaneously tracking the location of an agent. 
In interactive applications such as AR/VR and robotics, it is crucial for a SLAM system to possess not only accurate tracking and mapping capabilities but also \textit{global consistency} to ensure robustness and \textit{low latency} for optimal responsiveness.

   

Traditional SLAM systems, such as \cite{mur2015orb, mur2017orb, campos2021orb}, exhibit low latency, high-precision tracking, and employ loop detection to ensure global consistency. However, these systems are limited to constructing sparse point maps that lack dense geometry and texture information.


\begin{figure}[t]
\centering
\includegraphics[width=\linewidth,trim=4 4 4 4,clip]{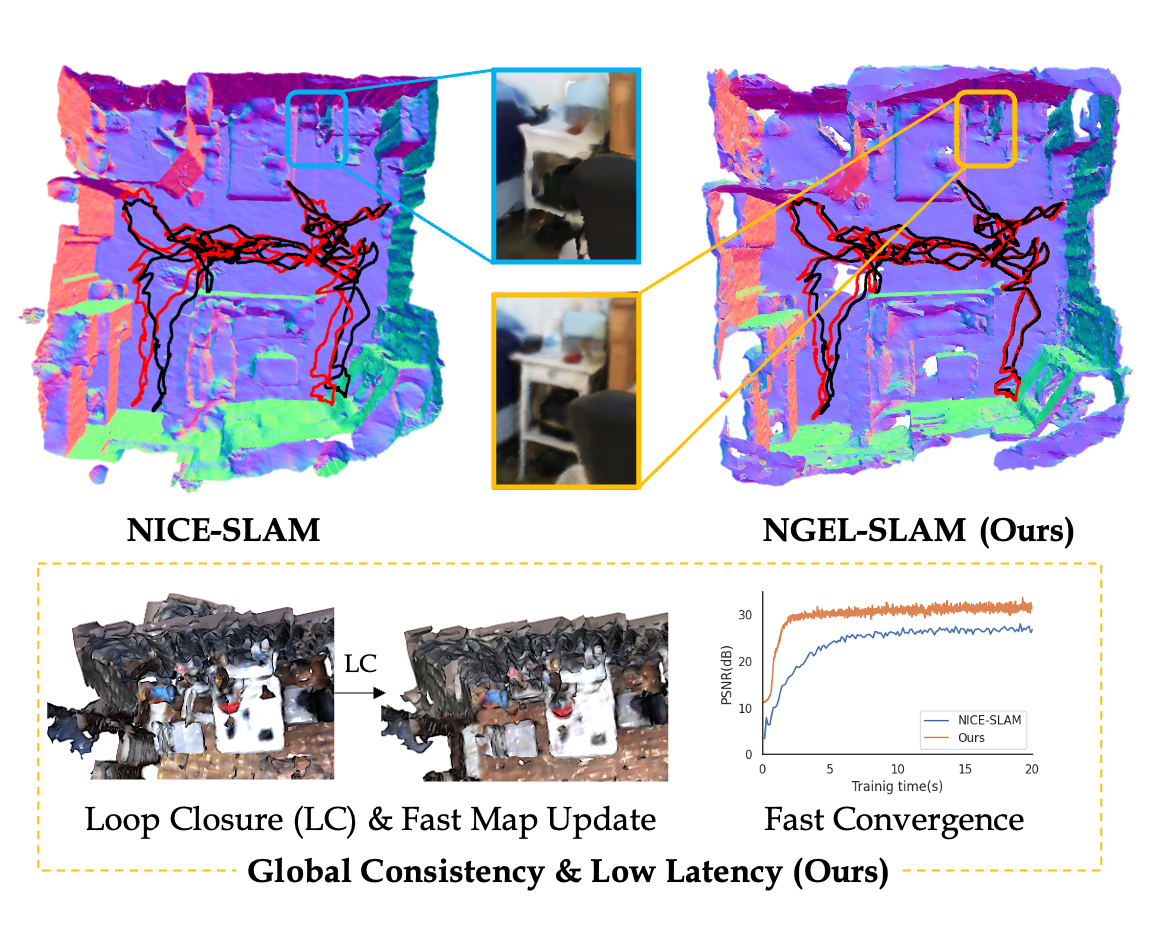}
       
\centering
\caption{\textbf{Rendering and tracking results.} Compared with NICE-SLAM \cite{zhu2022nice}, our method renders higher-fidelity images and provides more precise camera tracking results. 
Additionally, our method performs fast convergence and enables low-latency map updates after loop closure, enabling it to run 10x faster than NICE-SLAM.
The ground truth camera trajectory is shown in black, and the estimated trajectory is shown in red.
}
\label{normal mesh}
\end{figure}

Recent advances in neural implicit representations have enabled accurate and dense 3D surface reconstruction. Consequently, several neural implicit representation-based SLAM systems have been proposed. 
As pioneer works in this direction, iMAP \cite{sucar2021imap} and NICE-SLAM \cite{zhu2022nice} achieve both tracking and mapping based on the neural representation, resulting in high-fidelity scene reconstruction.
Nevertheless, the tracking based on neural representation lacks support for loop closure, leading to poor performance in large scenes due to the lack of global consistency, as illustrated in Fig.~\ref{normal mesh}.
Even if loop closure is integrated into their systems, e.g., by replacing tracking with traditional SLAM systems, re-training the entire map wrt. the updated pose is significantly time-consuming.
Moreover, the slow convergence of their mapping networks further prevents them from meeting the requirement of low-latency mapping.


To address these challenges, we propose NGEL-SLAM, a \textit{N}eural implicit representation-based \textit{G}lobal consist\textit{E}nt \textit{L}ow-latency SLAM system. 
NGEL-SLAM combines the tracking accuracy of the traditional SLAM system, ORB-SLAM3 \cite{campos2021orb}, with the capability of neural implicit representations to extract dense meshes and generate high-fidelity images. 
The traditional feature-based tracking module allows us to easily incorporate loop closure, enabling tracking with global consistency.
When a loop is detected, our mapping module immediately updates the map with low latency, thanks to the design of representing the scene as multiple neural implicit sub-maps. This avoids re-training the entire scene map upon loop closure but fixes most errors by simply updating the relative poses of the sub-maps. We further selectively fine-tune each sub-map based on the updated pose when necessary.
The mapping module updates the map every 14 ms and converges in a few iterations thanks to the octree-based sub-map representation, see Fig. \ref{normal mesh}.
The fast convergence and rapid response to loop closure make our system a truly low-latency system that achieves global consistency.
During inference, we introduce an uncertainty-based approach to select the best sub-maps for rendering an image at a given viewpoint.
We evaluate our method on a variety of real-world and synthetic datasets of different sizes and demonstrate its robustness and accuracy.

Our main contributions are summarized as follows:

\begin{itemize}
\item Accurate tracking and mapping. Our system achieves accurate performance by running the traditional feature-based tracking module and neural implicit scene mapping module in parallel.
\item Global consistency. Our system incorporates loop closure to ensure global consistency.
\item Low latency. Our system achieves both low latency tracking and mapping. The low latency mapping includes the quick response to loop closure and fast convergence given a new frame.
\item We conduct experiments on various datasets and demonstrate competitive performance compared to baselines.
\end{itemize}

\begin{figure*}[t]
\centering
\includegraphics[width=\linewidth,trim=4 4 4 4,clip]{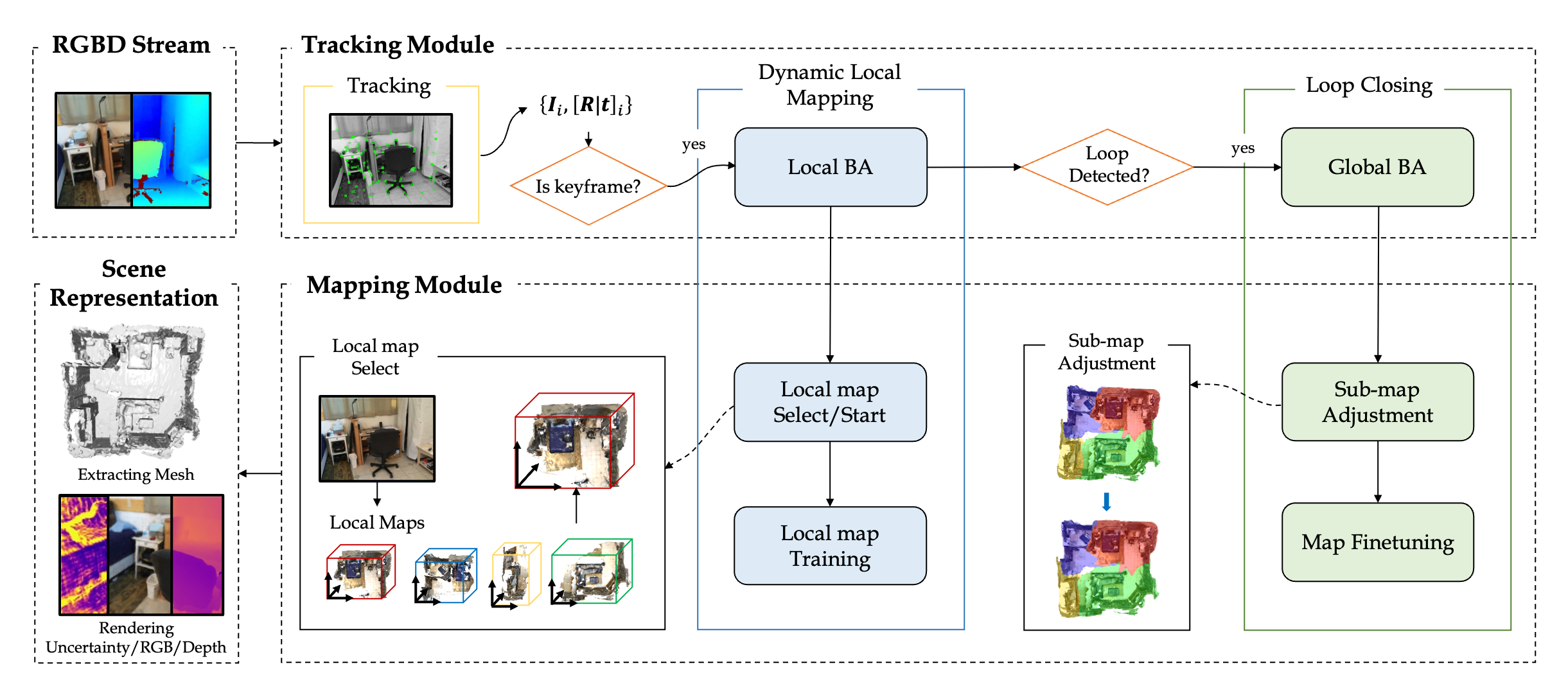}
\caption{\textbf{System overview.} Our proposed system comprises two primary modules: the tracking module and the mapping module. It can be further divided into three processes: tracking, dynamic local mapping, and loop closing.
These three processes work together to ensure global consistency and low latency in our system.
The tracking process takes an RGB-D stream as input and tracks the camera pose in real-time. If a frame is selected as a keyframe, it is passed to the dynamic local mapping process. In this process, the tracking module performs local BA while the mapping module trains the corresponding local map.
When a loop is detected, the loop closing process optimizes the camera poses using global BA and updates the scene representation. All processes are executed in parallel.
} 
\label{overview}
\end{figure*}


\section{Related Works}
\subsection{Visual SLAM}
Visual SLAM (Simultaneous Localization and Mapping) is an important research area in robotics and computer vision. Over the past two decades, significant advancements have been made in SLAM research.
Visual SLAM systems can be categorized into traditional SLAM systems \cite{davison2007monoslam, klein2007parallel, davison2007monoslam, engel2017direct, newcombe2011dtam, engel2014lsd, mur2015orb, mur2017orb, campos2021orb} and learning-based SLAM systems \cite{yokozuka2019vitamin, saputra2020deeptio, li2020deepslam}.
During the evolution of traditional SLAM systems, PTAM \cite{klein2007parallel} pioneers the separation of the SLAM task into tracking and mapping, achieving real-time performance.
A significant milestone in the advancement of visual SLAM was ORB-SLAM \cite{mur2015orb}, which employs an efficient feature-based approach to estimate camera trajectory and construct a 3D map. 
Traditional SLAM systems exhibit robustness to noise, allowing operation in diverse environments, scaling to large-scale environments, and handling of complex maps. However, they are limited to constructing sparse point-based maps while achieving accurate camera trajectory estimation.
In recent years, a line of works proposes to replace a part of the traditional SLAM system as a learning-based module, e.g., employing convolutional neural networks to extract features \cite{yokozuka2019vitamin, saputra2020deeptio, li2020deepslam}. These methods slow down the tracking process, while still facing the challenge of lacking a dense scene representation.
In our work, we take advantage of the great tracking performance of traditional SLAM systems, using ORB-SLAM3 \cite{campos2021orb} to track the camera poses and perform loop closure for maintaining global consistency.

\subsection{Neural Implicit Scene Representation}

Neural implicit representations have gained popularity in various applications, including 3D reconstruction \cite{park2019deepsdf, mescheder2019occupancy, sitzmann2020implicit, oechsle2021unisurf, chen2021mvsnerf,peng2020convolutional, wang2022go}, novel view synthesis \cite{mildenhall2021nerf, zhang2020nerf++, martin2021nerf} and 3D generative models \cite{niemeyer2021giraffe, schwarz2020graf, chan2021pi}.
In robotics, neural implicit representations have been used for object tracking and SLAM, enabling the construction of environment maps and estimation of robot or camera positions \cite{sucar2021imap, zhu2022nice, zhu2023nicer, johari2022eslam, yang2022vox, chung2023orbeez, haghighi2023neural}, which are closely related to our work. 
Among them, iMAP \cite{sucar2021imap}, NICE-SLAM \cite{zhu2022nice}, and ESLAM \cite{johari2022eslam}  achieve both tracking and mapping based on the neural representation.
However, in large-scale scenes, these methods suffer from the absence of loop closure and global bundle adjustment (BA), resulting in reduced tracking accuracy, a lack of global consistency, and inadequate training speed to meet low latency requirements.
While \cite{chung2023orbeez} and \cite{haghighi2023neural} enhance tracking accuracy by substituting neural tracking with a traditional method that incorporates loop closure, they fail to update the scene representation after loop closure, resulting in a lack of global consistency in their maps.
In this paper, we also employ the traditional tracking module but enable low-latency map update after loop closure, as we represent the scene with multiple local maps. While \cite{zhong2022shine, yu2023nf} also adopt the submap representation, they are offline systems. In contrast, we present a real-time SLAM system that consumes consecutive RGB-D frames and produces accurate poses and a global consistent map with low latency.

\section{System}
Fig.\ref{overview} provides an overview of our system. In this section, we introduce our system from the following aspects: tracking and mapping module (\ref{tmm}), dynamic local mapping (\ref{dlm}), loop closing (\ref{lc}), and uncertainty-based image rendering (\ref{render}).

\subsection{Tracking and Mapping Module} \label{tmm}
Our system achieves simultaneous estimation of accurate camera poses and 3D scene geometry and appearance from RGB-D video input through the utilization of two modules: tracking and mapping. The tracking module is based on ORB-SLAM3, the outstanding traditional SLAM system, and the mapping module represents the scene using multiple implicit neural maps.

The system comprises three processes: tracking, dynamic local mapping, and loop closing. 
During the tracking process, the tracking process estimates the camera pose $[\mathbf{R}|\mathbf{t}]_i$ and determines if the input frame $\mathbf{I}_i$ is a keyframe.
Each keyframe is fed into the dynamic local mapping process, which involves performing local BA by the tracking module and selecting the appropriate local map for training in the mapping module. 
Upon detecting a loop closure, the loop closing process begins, during which the tracking module optimizes all keyframe poses using global Bundle Adjustment (BA), and the mapping module promptly responds to significant changes in tracking poses by adjusting the sub-maps, followed by map fine-tuning.
All three processes run in parallel. The high-speed training and quick response to loop closure make our system meet the low latency requirement for real-world applications. Note that we refer to \textit{low latency} as the capability of integrating most of the information of a keyframe into the map before receiving a new keyframe from the captured RGB-D stream, regardless of whether a loop closure is detected or not.


\subsection{Dynamic Local Mapping} \label{dlm}
When a frame is decided to be a keyframe, the tracking module employs local BA to optimize the associated keyframes and provides the poses and the new keyframe to the mapping module. 
The mapping module first performs local map selection, in which it determines whether the new keyframe belongs to an existing local map by evaluating co-visible relations. This prevents the mapping module from generating redundant local maps, especially after loop closure.
In cases where the keyframe does not belong to any existing local map, a new local map is initialized and anchored relative to the current keyframe, known as the anchor frame.
After the local map is determined, the mapping module trains the local map with keyframes whose poses are optimized in local BA. The map is updated every 14ms, which ensures the low latency requirement.

\subsection{Loop Closing} \label{lc}

To ensure global consistency and rectify accumulated errors, loop detection is employed in our system.
When a loop is detected, the tracking module performs global BA.
However, global BA results in an immediate change in the previous predicted camera poses, commonly known as a trajectory jump. To address these changes, single volume-based implicit neural methods \cite{zhu2022nice, johari2022eslam} necessitate re-training of the scene representation and updates to most of the previously trained parameters, which is time-consuming.

In our system, we represent the entire scene with multiple local maps. When a global BA is completed, the scene representation undergoes a two-stage optimization from coarse to fine. In the first stage, the mapping module performs sub-map adjustment, which updates the scene representation by transforming the maps with the anchor keyframe poses. In the second stage, the mapping module fine-tunes the previous local maps to rectify errors. 
The first stage is a real-time adjustment that corrects errors among the local maps. Our experimental results demonstrate that this stage effectively rectifies a significant portion of the errors in the scene representation.
The second stage involves a sub-real-time optimization that eliminates small errors within the local maps, further enhancing the accuracy of the scene representation.


\subsection{Uncertainty-based Image Rendering} \label{render}
As our system incorporates multiple sub-maps, there are two scenarios to consider when rendering images from a given viewpoint. The first scenario arises when the view frustum intersects a sub-map entirely, allowing us to render the image using that specific sub-map. The second scenario occurs when the view frustum lies at the boundary of different sub-maps, making it impossible to generate a complete image from a single sub-map. In such cases, we employ a pixel-wise fusion of the image based on the lowest uncertainty.

\begin{figure}[t]
\centering
\includegraphics[width=\linewidth,trim=4 4 4 4,clip]{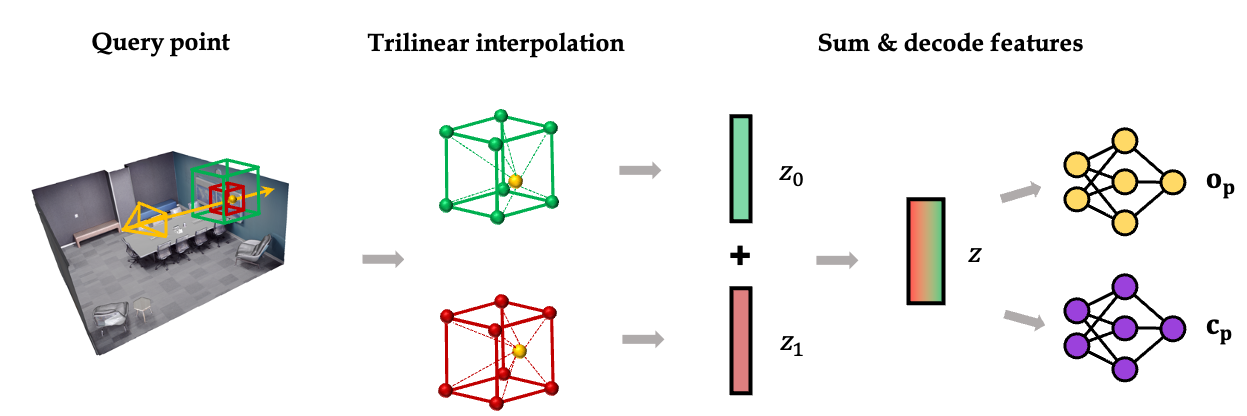}
\caption{\textbf{Mapping network.} The Mapping network employs a sparse octree structure to store multi-level features and two single MLPs. 
} 
\label{network}
\end{figure}

\section{Map Representation and Training}

\subsection{Octree-based Implicit Neural Representation} \label{framework}

Voxel grid-based NeRF architectures \cite{zhu2022nice, wang2022go} have been shown to converge quickly. However, since most of the space is unoccupied, the dense grid structure used in these methods is memory-wasting. 
To address this issue, we draw inspiration from NGLOD \cite{takikawa2021neural} and use a sparse octree-based grid. The octree only grows where the space is occupied, making it memory-efficient. 
Multi-level feature vectors, denoted by $\mathbf{z}$, are stored on the nodes of the octree. 
As shown in Fig. \ref{network}, when we query a point $\mathbf{p}$, the feature $\mathbf{z}^{(i)}_\mathbf{p}$ of $\mathbf{p}$ in level $i$ is obtained by trilinear interpolation. 
Then, similar to NGLOD, we sum the features in different active levels (i.e., levels that store features) to obtain $\mathbf{z_p}$ as the feature of $\mathbf{p}$.
Furthermore, we have modified the octree to an incremental paradigm, where a Morton code table is maintained to provide the position of new points added to the octree.

In our method, we employ two small MLP decoders, one for occupancy and another for color. To calculate the occupancy $o_\mathbf{p}$ and color $\mathbf{c}_\mathbf{p}$ at a given point $\mathbf{p}\in\mathbb{R}^3$ in space, we use the following equations:

\begin{equation}
o_\mathbf{p} = \sigma(f_{occ}(\mathbf{z}_\mathbf{p})), \quad \mathbf{c}_\mathbf{p} = f_{color}(\mathbf{z}_\mathbf{p})
\end{equation}

Here, $\mathbf{z_p}$ is the feature vector at point $\mathbf{p}$, $f_{occ}$ and $f_{color}$ are the occupancy and color decoders, respectively, and $\sigma$ is the sigmoid function. 

\begin{table*}[t]
\centering
\caption{\textbf{Quantitative comparison of mapping on the Replica dataset.} Data averaged from eight scenes. \texttt{GT pose} and \texttt{Est pose} respectively represent rendering with ground truth poses and rendering with estimated poses.}
\begin{tabular}{lccc|ccc}
\toprule
 \makebox[0.12\textwidth] & \multicolumn{3}{c|}{\texttt{Evaluation @ GT Pose}} & \multicolumn{3}{c} {\texttt{Evaluation @ Est Pose}}\\
 \midrule
       & \makebox[0.12\textwidth]{NICE-SLAM \cite{zhu2022nice}} & \makebox[0.12\textwidth]{ESLAM \cite{johari2022eslam}}  & \makebox[0.12\textwidth]{\textbf{Ours}} & \makebox[0.12\textwidth]{NICE-SLAM \cite{zhu2022nice}} & \makebox[0.12\textwidth]{ESLAM \cite{johari2022eslam}}  & \makebox[0.12\textwidth]{\textbf{Ours}}\\
\midrule
Depth L1 [cm] $\downarrow$ & 3.53 & 2.81 & \textbf{1.28}& 1.58 & 2.28 & \textbf{0.65}\\
PSNR $\uparrow$ & 21.98 & 27.33  & \textbf{29.53}& 26.20  & 29.25  & \textbf{30.44}\\
SSIM $\uparrow$ & 0.775 & 0.836   & \textbf{0.864} & 0.832   & 0.855   & \textbf{0.876}\\
LPIPS $\downarrow$  & 0.247 & 0.204 & \textbf{0.156}& 0.233  & 0.190 & \textbf{0.151}\\
\bottomrule
\end{tabular}
\label{recon_replica}
\end{table*}

\subsection{Volume Rendering} \label{rendering}
To optimize our scene representation framework in section \ref{framework}, we use differentiable volume rendering proposed in NeRF \cite{mildenhall2021nerf}. Given camera intrinsic parameters and current camera pose, we cast a ray $\mathbf{r}$ from the camera center $\mathbf{o}$ through the pixel along its normalized view direction $\mathbf{v}$.

\subsubsection{Depth and Color}We sample $N$ points along a ray, denoted as $\mathbf{x}_i = \mathbf{o} + d_i\mathbf{v}$, where $d_i$ is the depth of point $\mathbf{x}_i$, and $i \in {1,\cdots, N}$. The predicted occupancy values and color values of these points are denoted as $o_i$ and $\mathbf{c}_i$, respectively. For a given ray $\mathbf{r}$, we can calculate the depth $\hat{D}$ and color $\hat{\mathbf{C}}$ as:

\begin{equation}
\hat{D}(\mathbf{r}) = \sum_{i=1}^N T_i\alpha_i d_i \quad \text{and} \quad \hat{\mathbf{C}}(\mathbf{r}) = \sum_{i=1}^N T_i\alpha_i \mathbf{c}_i
\label{dc}
\end{equation}

Here, $\alpha_i = o_i$ and $T_i = \prod_{j=1}^i(1-o_j)$ correspond to the transmittance and alpha value of sample point $i$ along the ray $\mathbf{r}$, respectively.

\subsubsection{Voxel-based sampling}
To fully utilize the octree structure, we adopt a voxel-based sampling policy. Given a ray $\mathbf{r}$, we query the voxels it intersects and sample points along the ray within these voxels. Since the octree only grows where the points are occupied, the sampled points are all either close to the object surface or inside the objects. We sample a fixed number of $N_{point}$ points per voxel. Therefore, for a ray $\mathbf{r}$ intersecting with $N_{voxel}$ voxels, we sample $N=N_{point}\times N_{voxel}$ points along the ray.

\subsubsection{Optimizing}
To optimize the scene representation features and decoders in Section \ref{framework}, we uniformly select $M$ pixels in the current frame and train the scene representation using photometric loss and geometry loss. The photometric loss is the mean squared error (MSE) loss between the rendered and ground truth color images, while the geometry loss is a simple $\mathcal{L}_1$ loss between the rendered and ground truth depths. Specifically, we define the losses as follows:

\begin{equation}
\mathcal{L}_p = \frac{1}{M}\sum_{i=1}^{M}|\hat{\textbf{C}}-\textbf{C}_{gt}|_2
\quad
\mathcal{L}_g = \frac{1}{M}\sum_{i=1}^{M}|\hat{D}-D_{gt}|
\end{equation}

We optimize the features $\mathbf{z}$ and decoder parameters $\theta$ jointly by minimizing the loss function:

\begin{equation}
\min_{\mathbf{z}, \theta}\quad \lambda_p\mathcal{L}_p + \mathcal{L}_g
\end{equation}

Here, $\lambda_p$ is the weight of the photometric loss.

\subsubsection{Uncertainty}
Since occupancy follows a Bernoulli distribution, the occupancy value $o_\mathbf{p}$ of a point $\mathbf{p}$ represents the probability that the point is occupied, and its variance can be calculated as $Var = o_\mathbf{p}(1-o_\mathbf{p})$. The occupancy variance of a ray can then be rendered as:
\begin{equation}
    \sigma_{o}^2(\textbf{r}) = \frac{1}{N}\sum_{i=1}^N o_i(1-o_i)
\end{equation}

To account for the higher uncertainty in unobserved regions, we set the variance of these regions to be $0.25$.

\section{Experiment}
In this section, we validate our SLAM framework on real and synthetic datasets of varying sizes and complexities. We also compare our method with existing implicit representation-based methods.

\begin{figure}
\centering
\captionsetup[subfloat]{labelfont=scriptsize,textfont=scriptsize}
    \subfloat[NICE-SLAM]{
    \centering
        \begin{minipage}{0.23\linewidth}
            \centering
            \includegraphics[width=\linewidth]{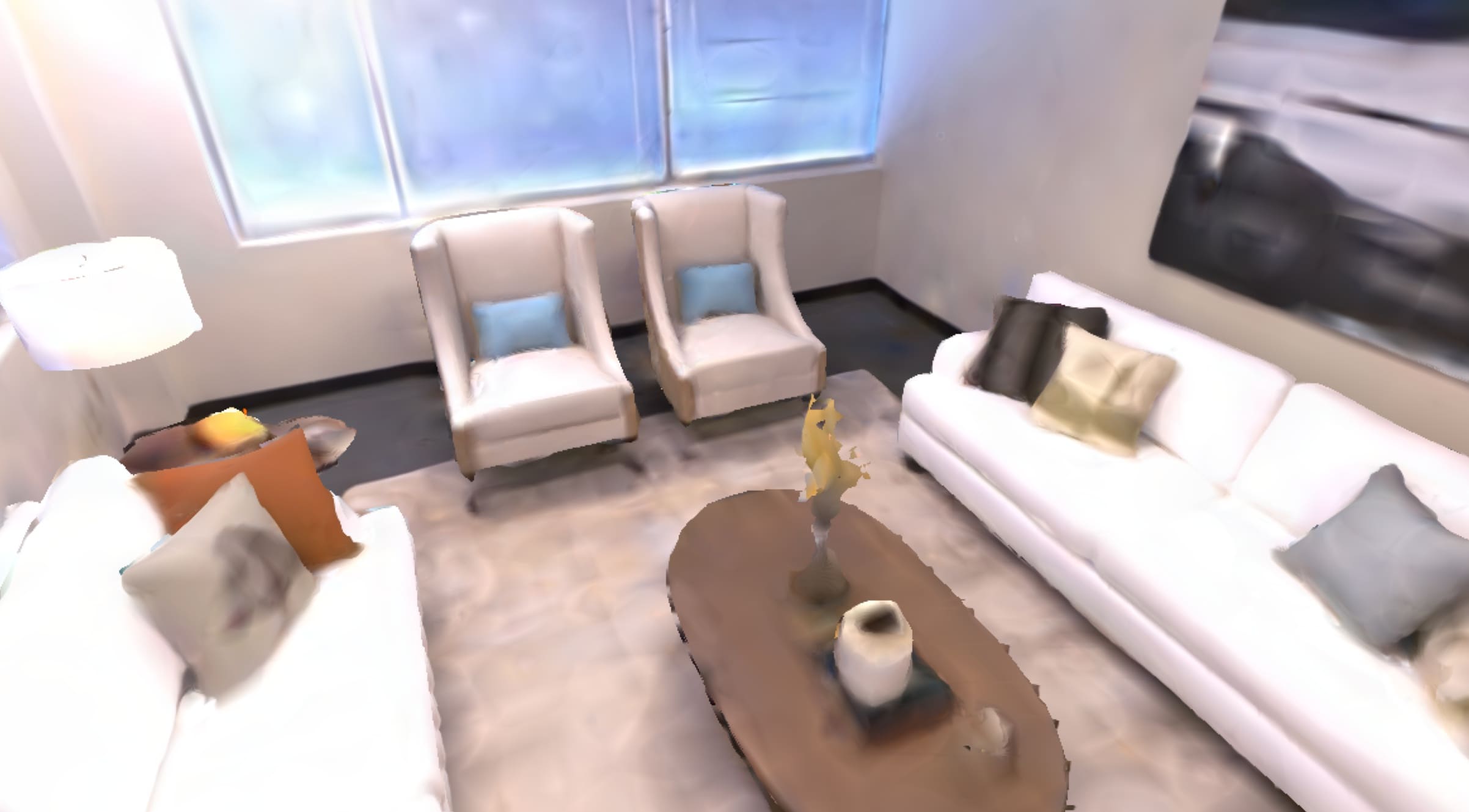}\vspace{1mm}
            \includegraphics[width=\linewidth]{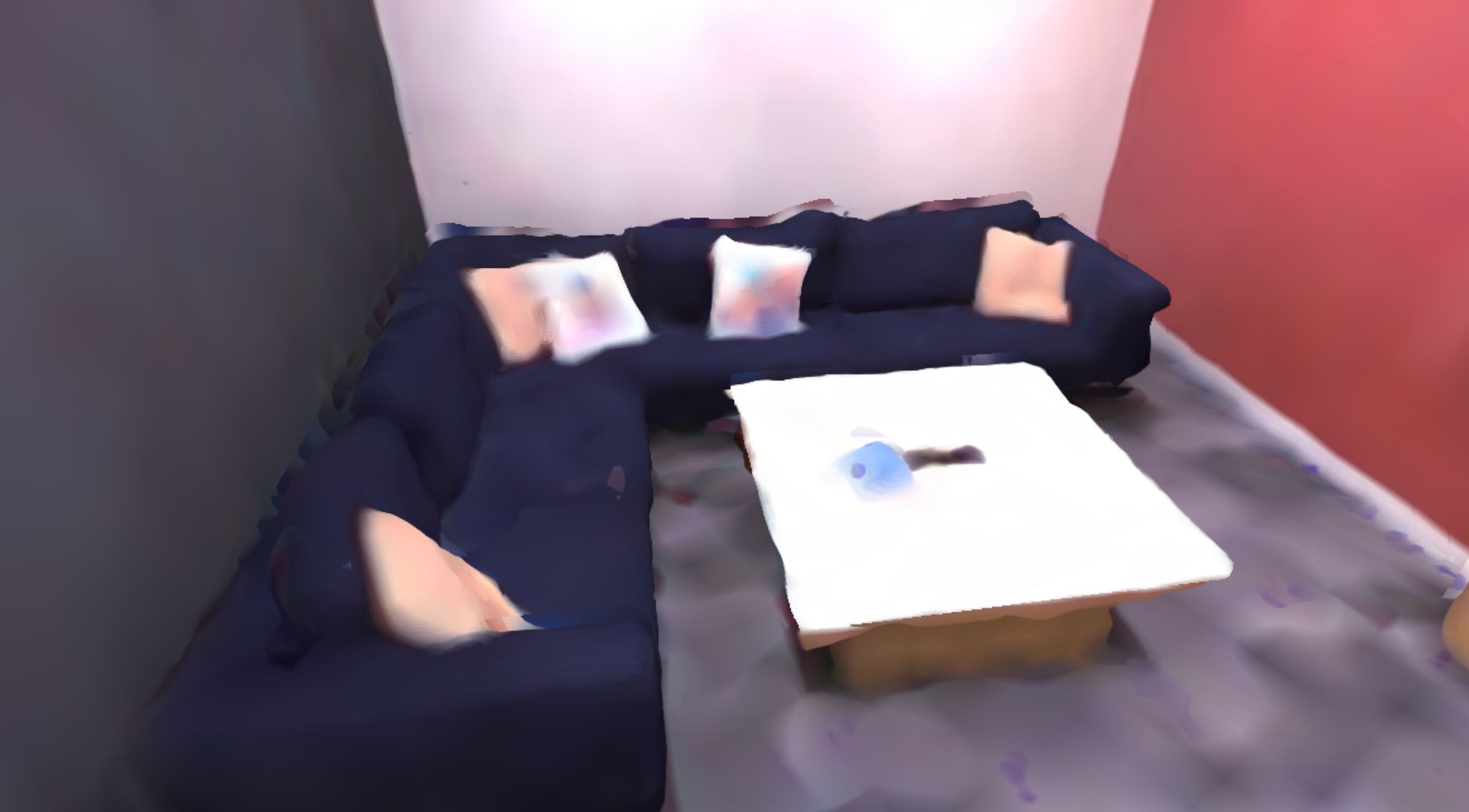}
        \end{minipage}
        }\hspace{-3mm}
    \subfloat[ESLAM]{
    \centering
        \begin{minipage}{0.23\linewidth}
            \centering
            \includegraphics[width=\linewidth]{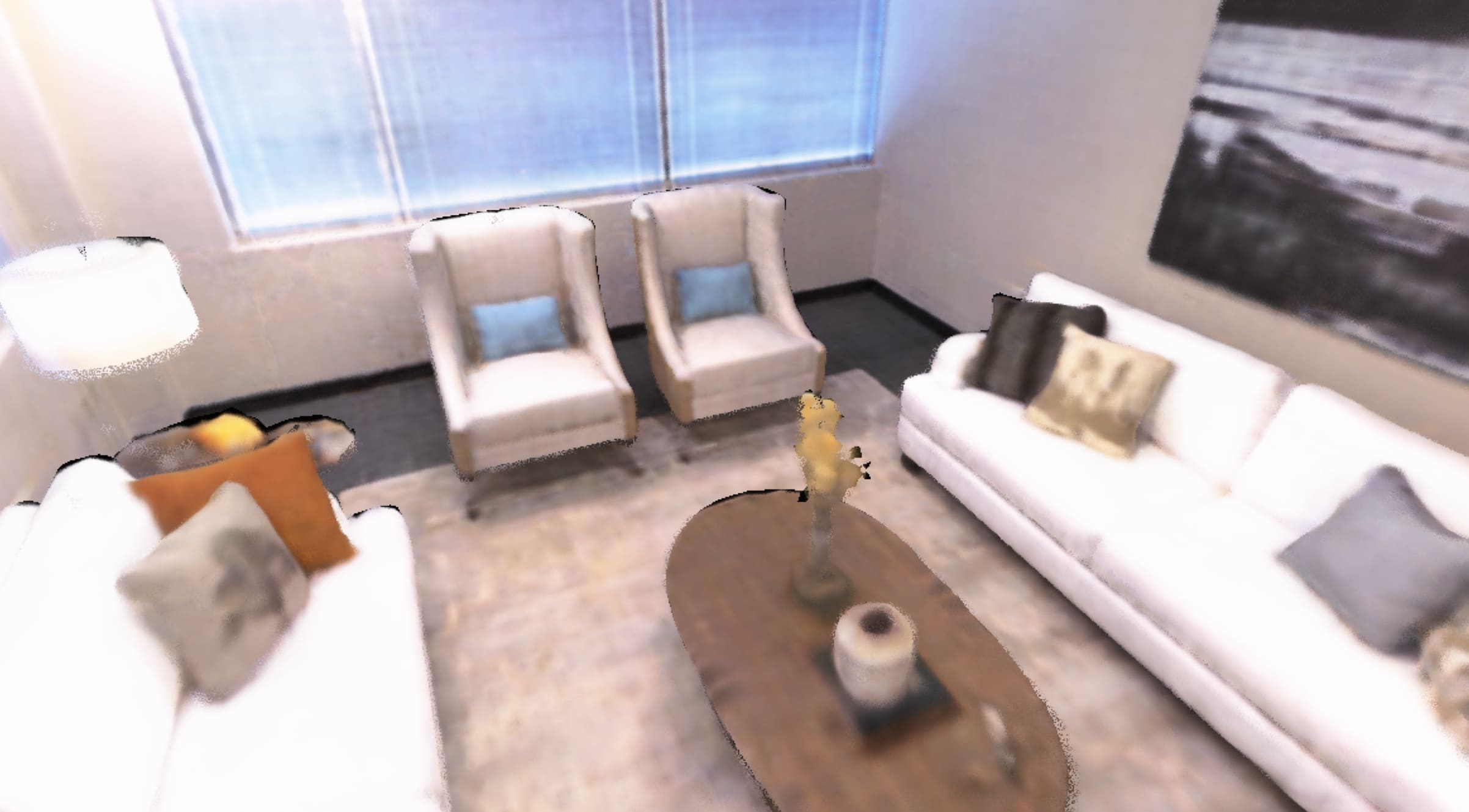}\vspace{1mm}
            \includegraphics[width=\linewidth]{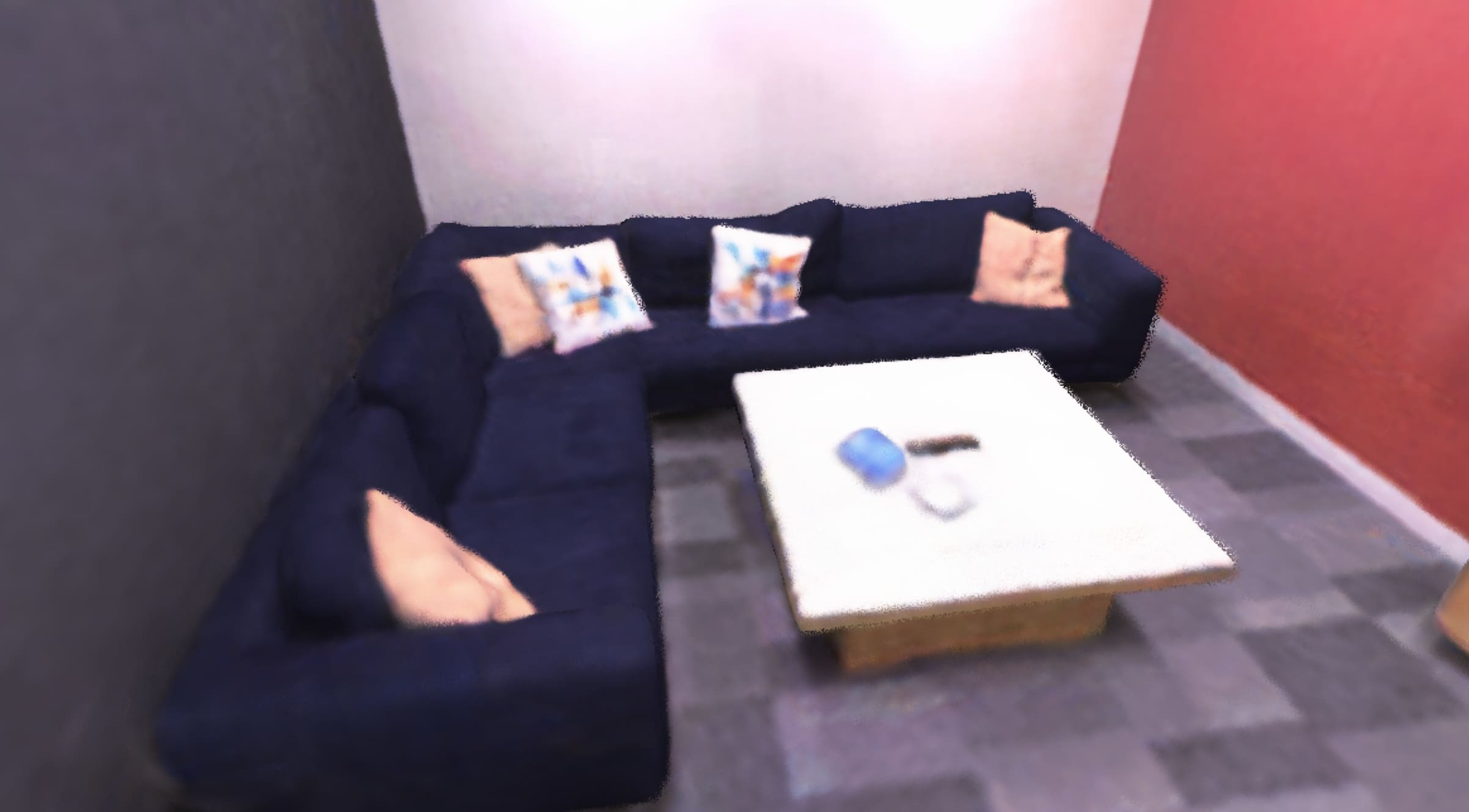}
        \end{minipage}
        }\hspace{-3mm}
    \subfloat[\textbf{Ours}]{
    \centering
        \begin{minipage}{0.23\linewidth}
            \centering
            \includegraphics[width=\linewidth]{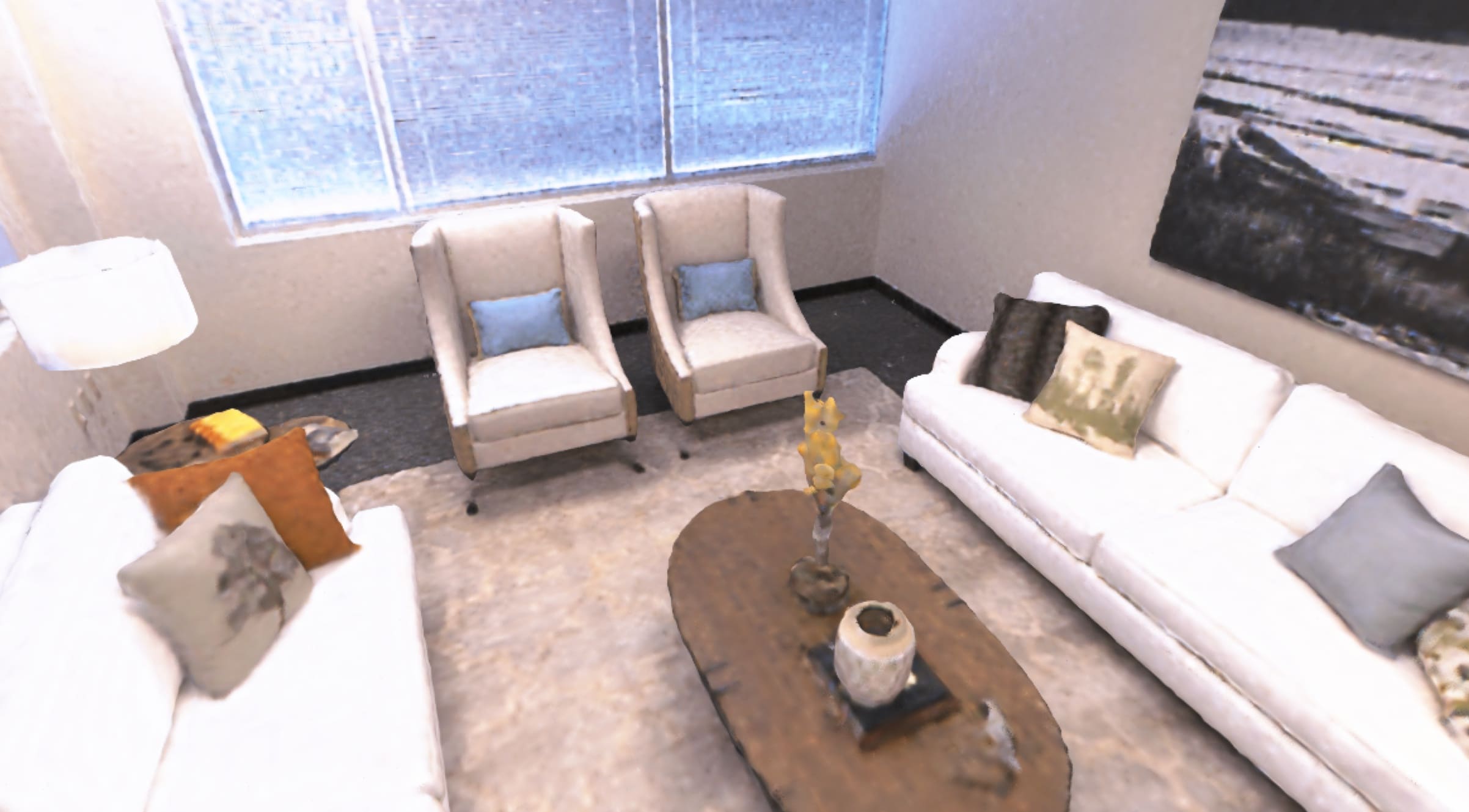}\vspace{1mm}
            \includegraphics[width=\linewidth]{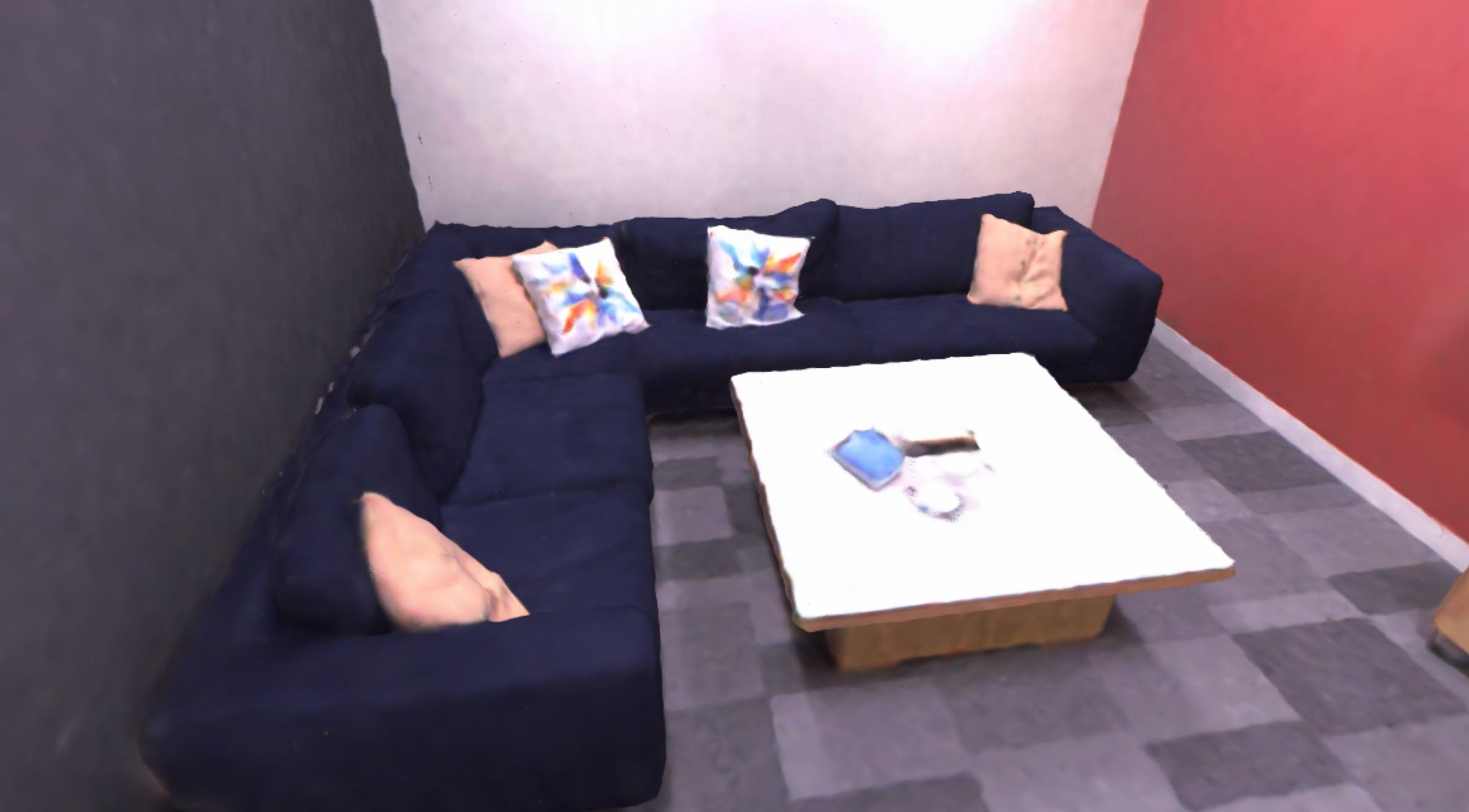}
        \end{minipage}
        }\hspace{-3mm}
    \subfloat[GT]{
    \centering
        \begin{minipage}{0.23\linewidth}
            \centering
            \includegraphics[width=\linewidth]{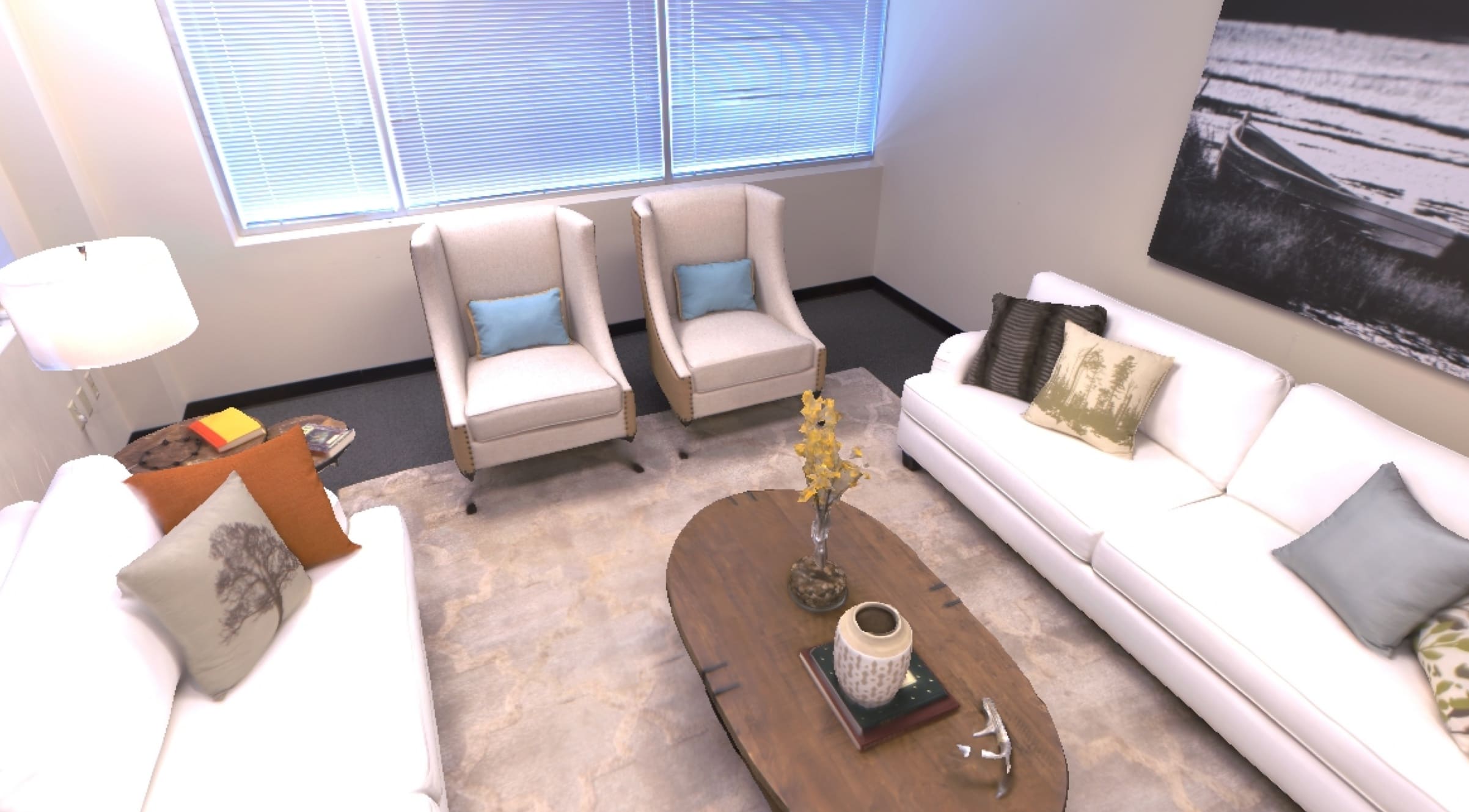}\vspace{1mm}
            \includegraphics[width=\linewidth]{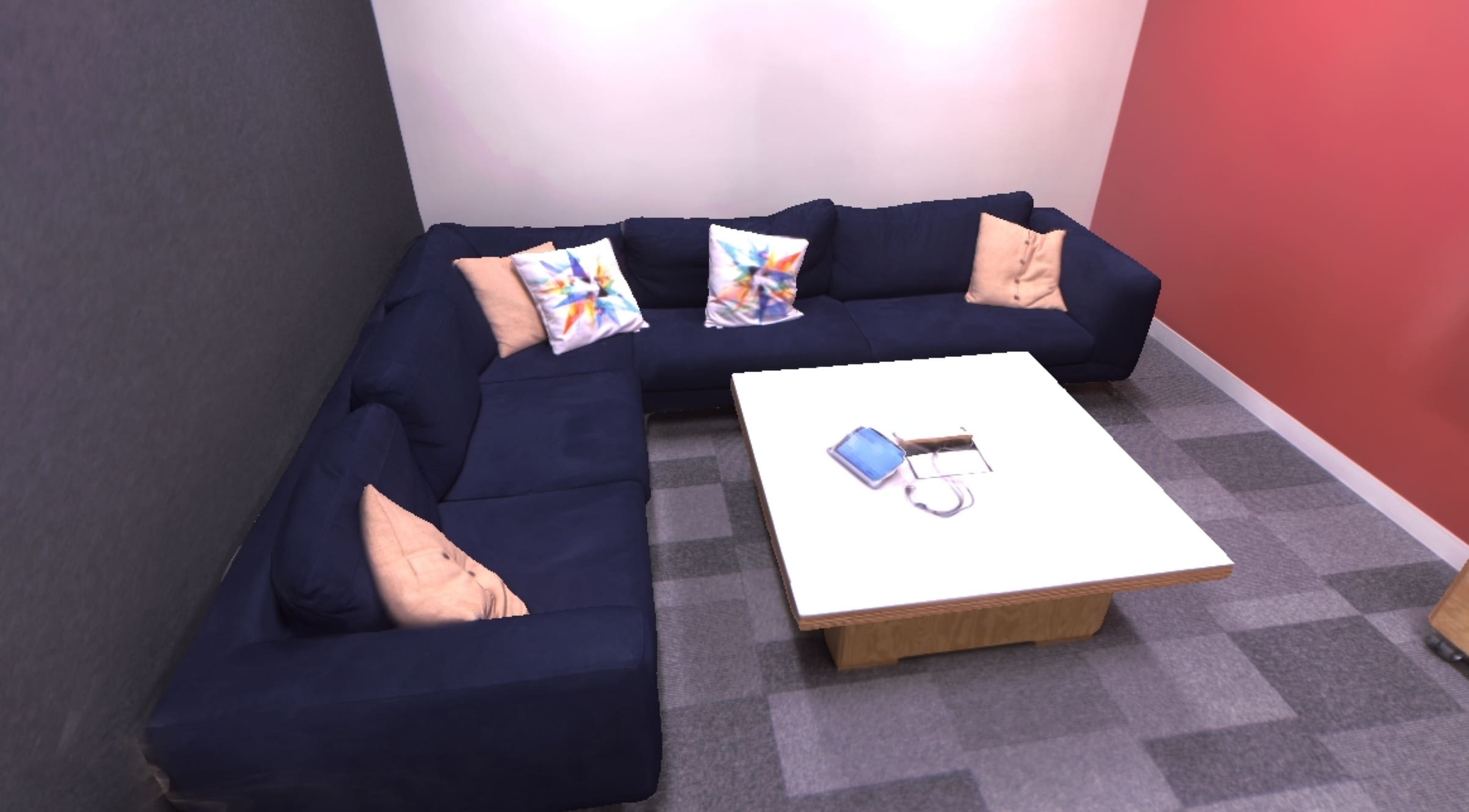}
        \end{minipage}
        }
        
\centering
\caption{\textbf{Rendering results on the Replica dataset.} 
}
\label{replica_color}
\end{figure}

\subsection{Experiment Setup}

\subsubsection{Datasets}
We consider 4 datasets for evaluation: Replica \cite{replica19arxiv}, ScanNet \cite{dai2017scannet}, TUM RGB-D dataset \cite{sturm12iros} for tracking and mapping capability analysis, and the apartment dataset provided by NICE-SLAM for ablation study. 

\subsubsection{Baselines}
We compare our method to three existing open-source state-of-the-art RGB-D implicit neural SLAM systems: iMAP \cite{sucar2021imap}, NICE-SLAM \cite{zhu2022nice}, and ESLAM \cite{johari2022eslam}. The experiments are designed following NICE-SLAM \cite{zhu2022nice}.

\subsubsection{Metrics}
To evaluate the geometry and appearance, we use L1 loss for rendered depth images and the PSNR, SSIM, and LPIPS for rendered color images on 100 randomly sampled poses. Additionally, we evaluate camera tracking using ATE RMSE.

\subsubsection{Implementation Details}

Our SLAM system runs on a single NVIDIA RTX 3090 GPU. 
We set the number of sampling points along a ray in every intersecting voxel to $N_{point}=10$ and photometric loss weighting to $\lambda_p=1$. We use sample $M=5000$ pixels. Decoders are both MLPs with a hidden feature dimension of 32 and 2 fully-connected blocks. The RGBD stream is input at a rate of 10Hz.

\begin{table}[t]
\centering
\caption{\textbf{Quantitative comparison of tracking on TUM RGB-D.} ATE-RMSE [cm] is used as the metric.}

\begin{tabular}{lccc}
\toprule
        \makebox[0.1\textwidth]& \makebox[0.08\textwidth]{\texttt{fr1/desk}} & \makebox[0.08\textwidth]{\texttt{fr2/xyz}} & \makebox[0.08\textwidth]{\texttt{fr3/office}}\\
\midrule
iMAP \cite{sucar2021imap} & 4.9 & 2.0 & 5.8\\
NICE-SLAM \cite{zhu2022nice} & 2.7 & 1.8 & 3.0  \\
ESLAM \cite{johari2022eslam} & 2.5 & 1.1 & 2.4 \\
\textbf{Ours}  & \textbf{1.5} & \textbf{0.5}  & \textbf{1.0} \\
\bottomrule
\end{tabular}
\label{tum_tracking}
\end{table}

\subsection{Experiment Results}
\subsubsection{Evaluation on Replica \cite{replica19arxiv}}
We evaluate our method's scene representation capabilities on 8 scenes from the Replica dataset and compare its performance in geometry and appearance representation to other baseline methods. 
Following other papers, we first evaluate the reconstruction performance by rendering the images with ground truth camera poses, which means the results include tracking errors.
Furthermore, to better independently verify model reconstruction capabilities, we use the camera poses estimated by the tracking modules to render the images and evaluate the results.
The results presented in TABLE \ref{recon_replica} demonstrate that our method outperforms baseline methods significantly across all metrics. 
Qualitatively, as shown in Fig. \ref{replica_color}, our method produces geometry that is more accurate and detailed, with higher-fidelity textures.

\begin{table}[t]
\centering
\caption{\textbf{Quantitative comparison of tracking on ScanNet.} ATE-RMSE [cm] is used as the metric.}
\setlength{\tabcolsep}{1.2mm}{
\begin{tabular}{lccccccc}
\toprule
        \texttt{Scene ID} & \texttt{0000} & \texttt{0059} & \texttt{0106}
        & \texttt{0169} & \texttt{0181} & \texttt{0207} & Avg.\\
\midrule
iMAP \cite{sucar2021imap} & 55.95 & 32.0 & 17.50 & 70.51& 32.10 & 11.91 & 36.67 \\
NICE-SLAM \cite{zhu2022nice} & 8.64 & 12.25 & 8.09 & 10.28 & 12.93 & \textbf{5.59} & 9.63  \\
ESLAM \cite{johari2022eslam} & 7.41 & 8.64 & \textbf{7.48} & 6.73 & \textbf{9.03} & 5.72 & 7.50 \\
\textbf{Ours}  & \textbf{7.23} & \textbf{6.98}  & 7.95 & \textbf{6.12}& 10.14& 6.27 & \textbf{7.44}\\
\bottomrule
\end{tabular}
}
\label{scannet_tracking}
\end{table}

\begin{figure}[t]
\centering
\captionsetup[subfloat]{labelfont=scriptsize,textfont=scriptsize}
    \subfloat[NICE-SLAM]{
    \centering
        \begin{minipage}{0.23\linewidth}
            \centering
            \includegraphics[width=\linewidth]{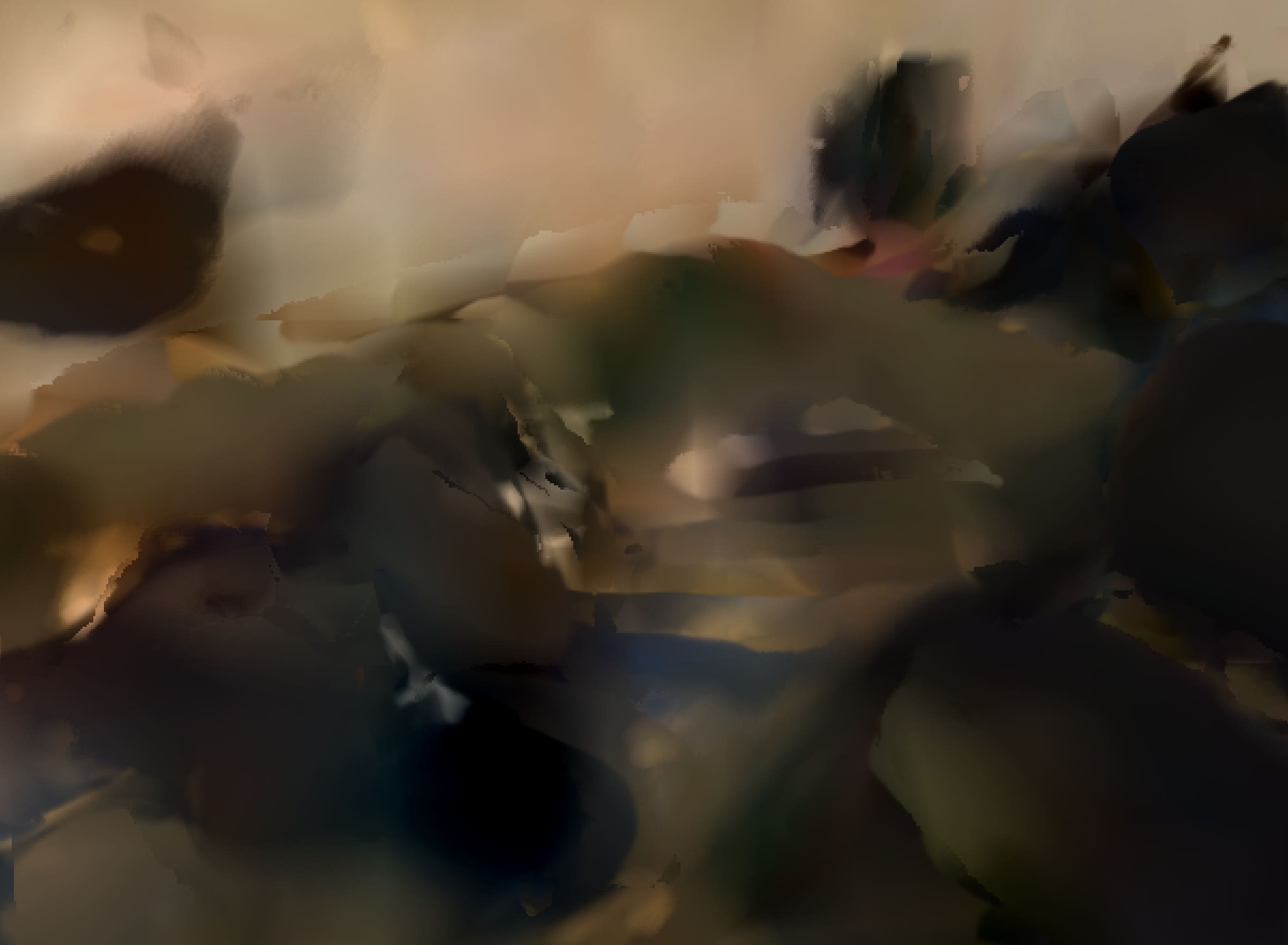}\vspace{1mm}
            \includegraphics[width=\linewidth]{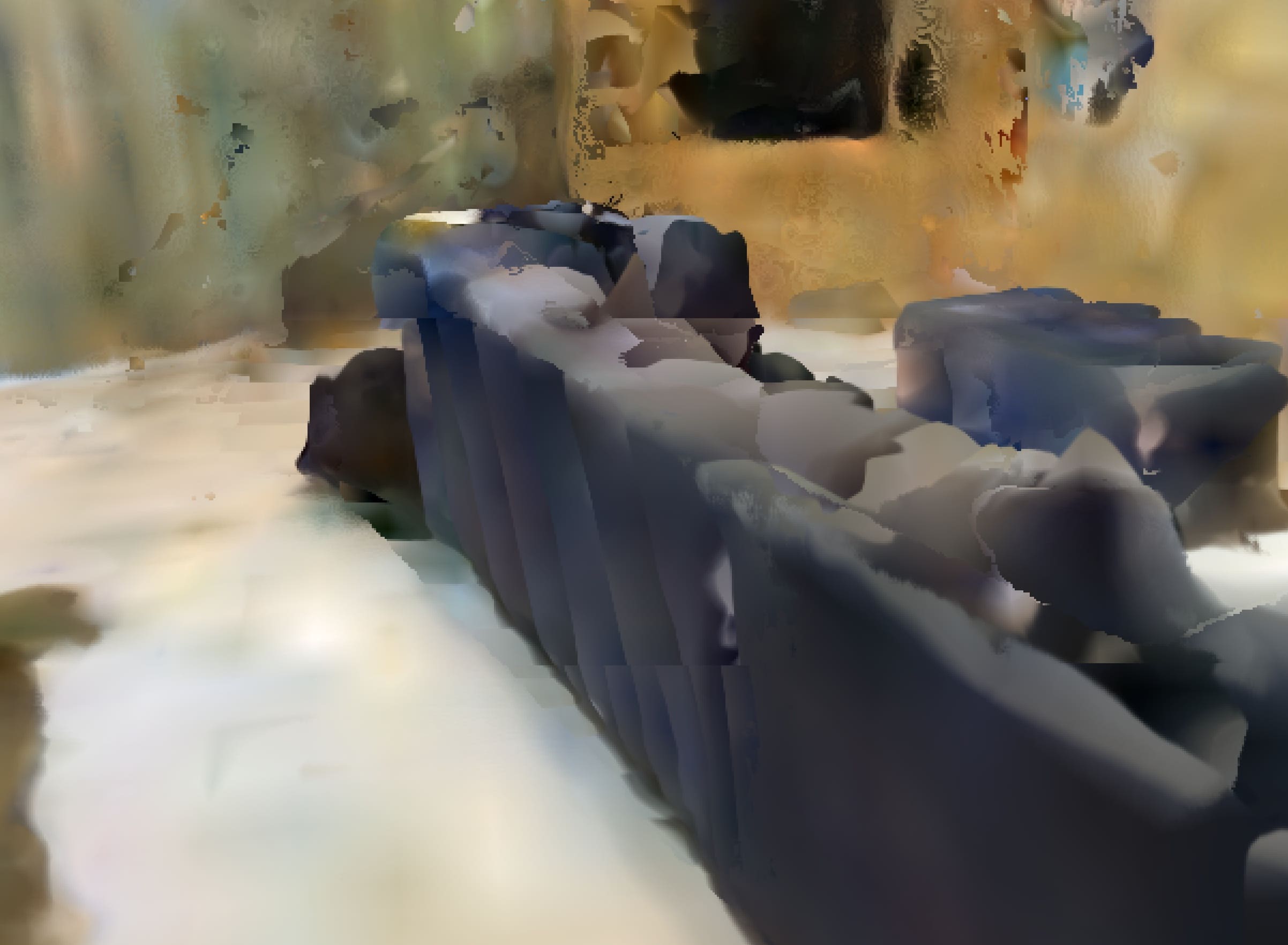}
        \end{minipage}
    }\hspace{-3mm}
    \subfloat[ESLAM]{
    \centering
        \begin{minipage}{0.23\linewidth}
            \centering
            \includegraphics[width=\linewidth]{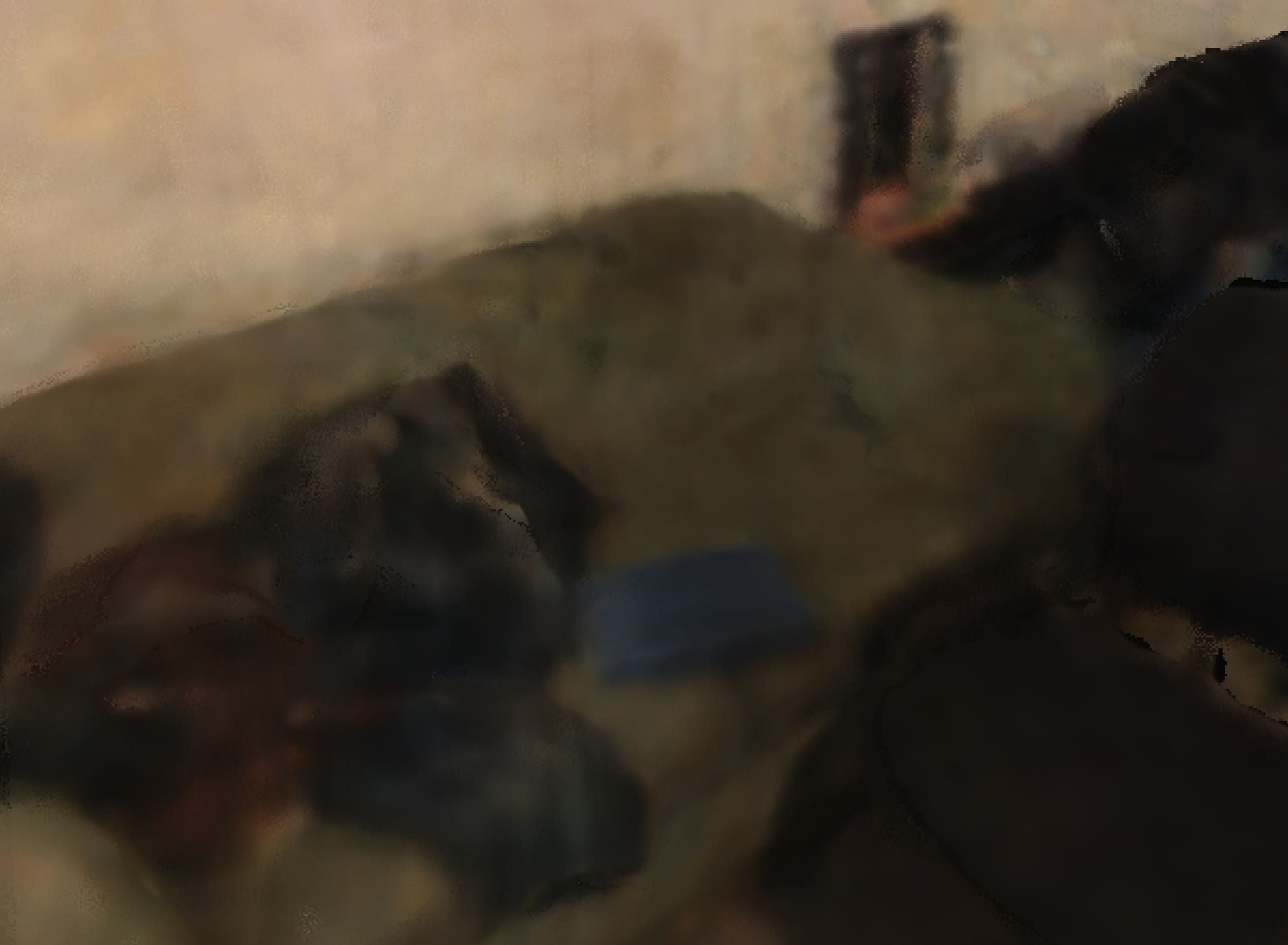}\vspace{1mm}
            \includegraphics[width=\linewidth]{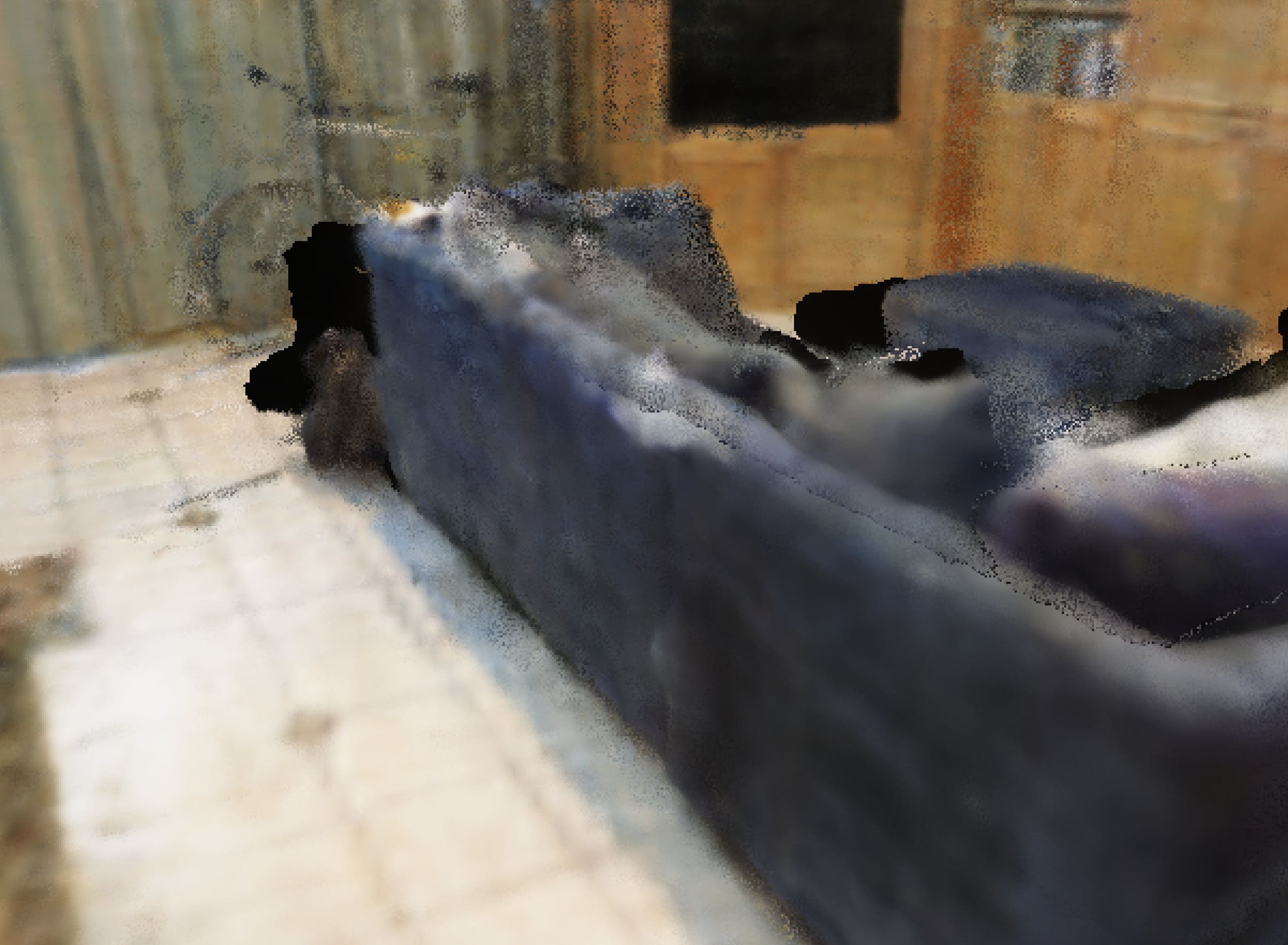}
        \end{minipage}
    }\hspace{-3mm}
    \subfloat[\textbf{Ours}]{
    \centering
        \begin{minipage}{0.23\linewidth}
            \centering
            \includegraphics[width=\linewidth]{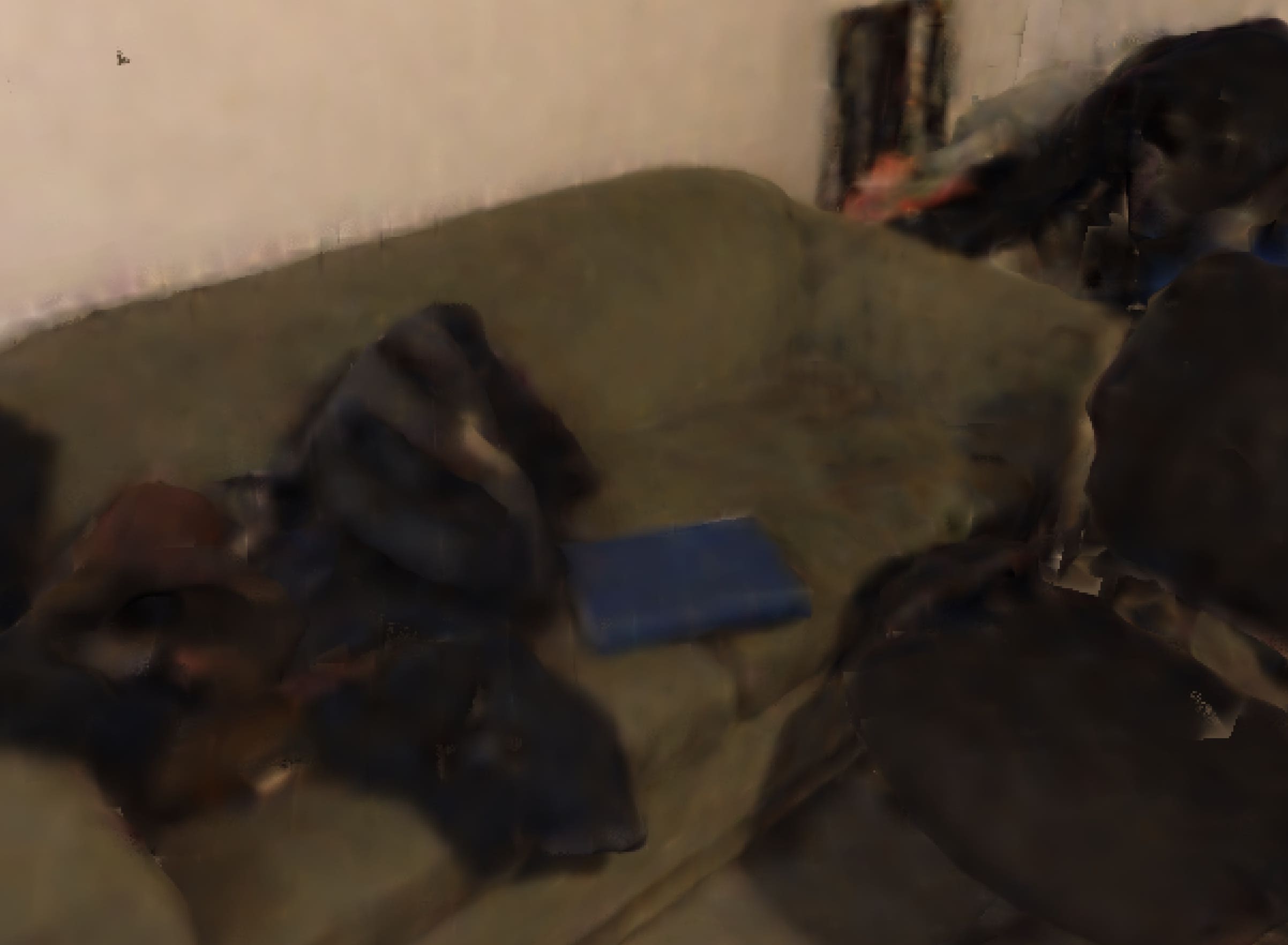}\vspace{1mm}
           \includegraphics[width=\linewidth]{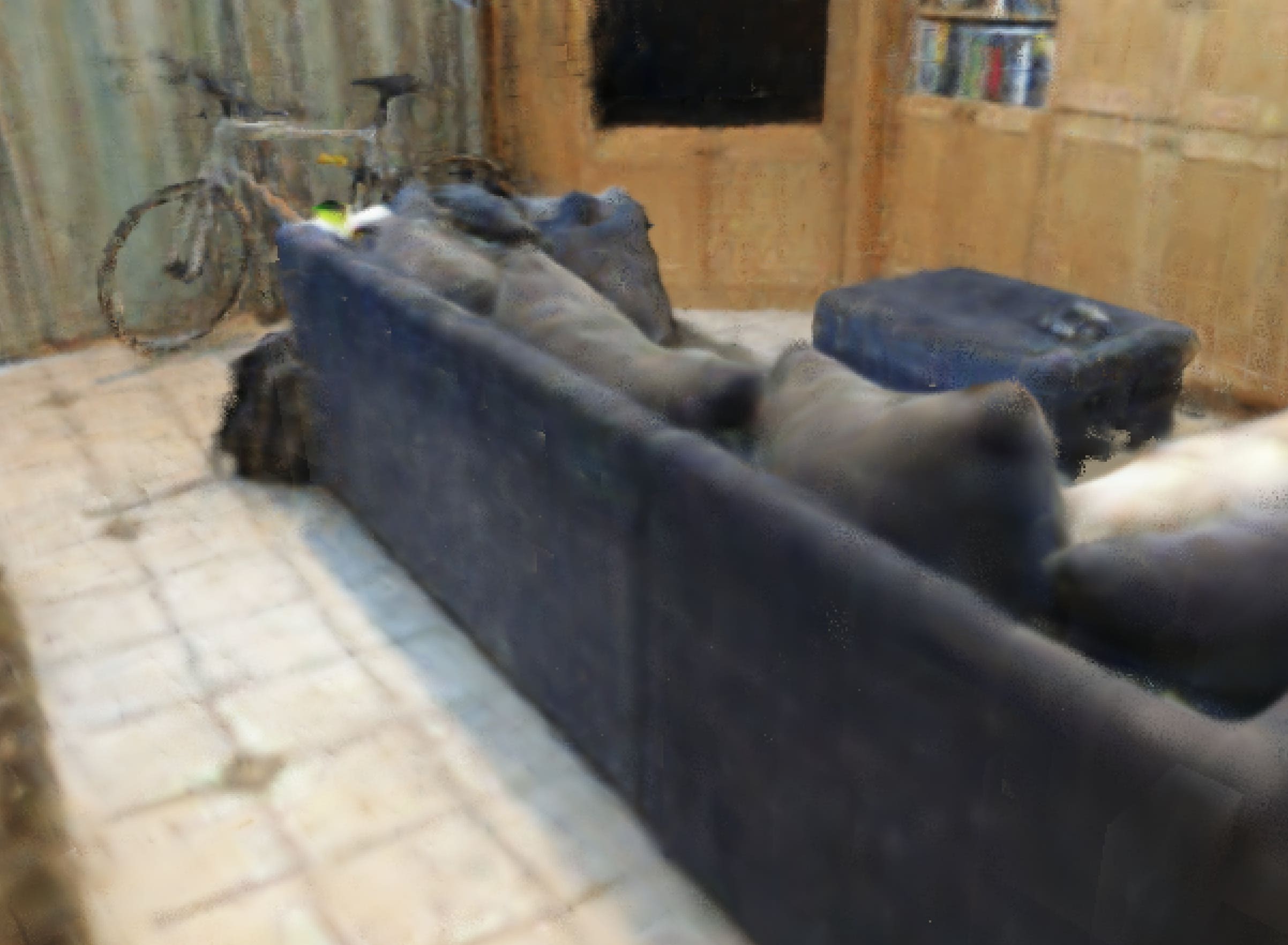}
        \end{minipage}
    }\hspace{-3mm}
    \subfloat[GT]{
    \centering
        \begin{minipage}{0.23\linewidth}
            \centering
            \includegraphics[width=\linewidth]{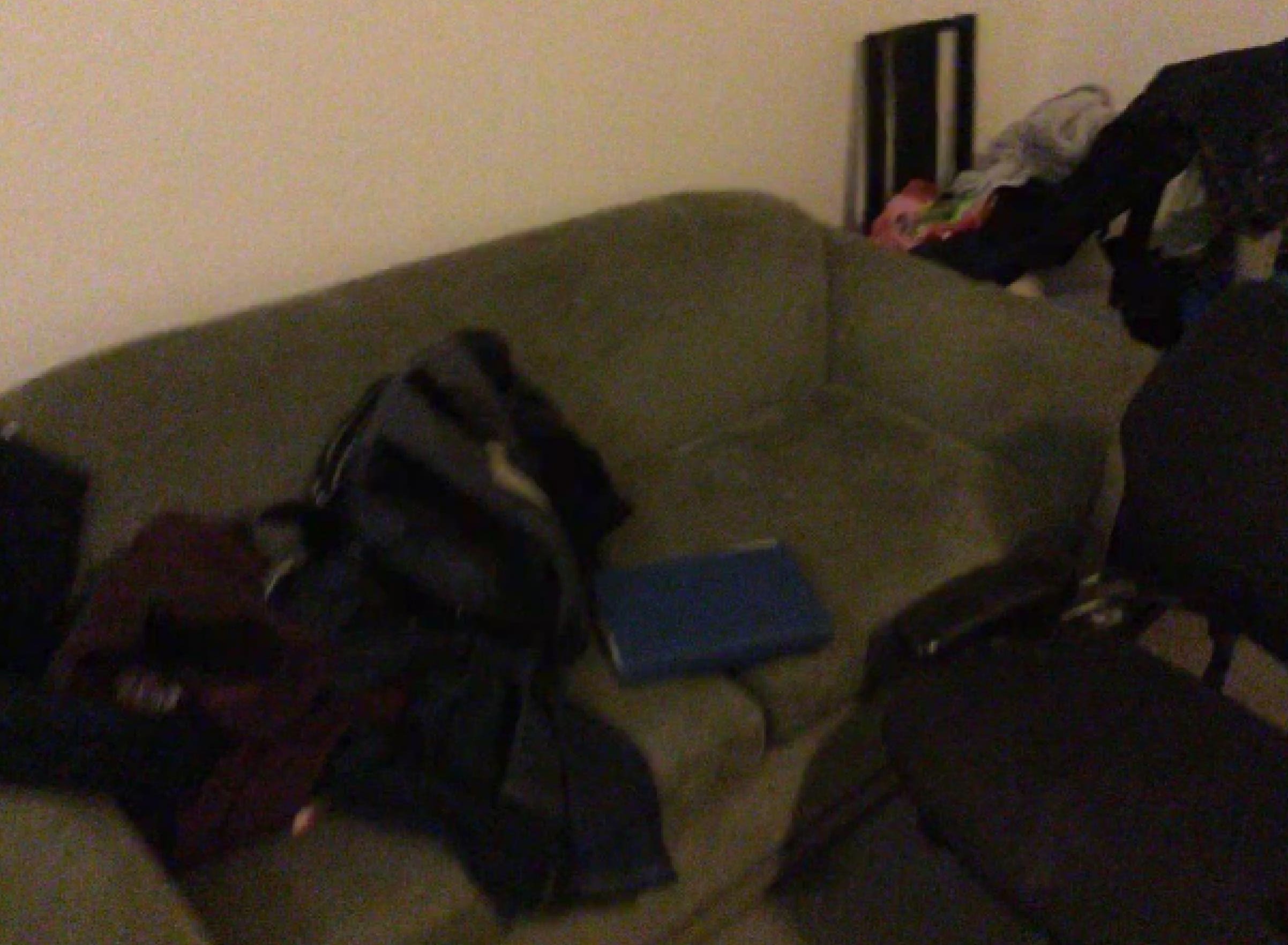}\vspace{1mm}
            \includegraphics[width=\linewidth]{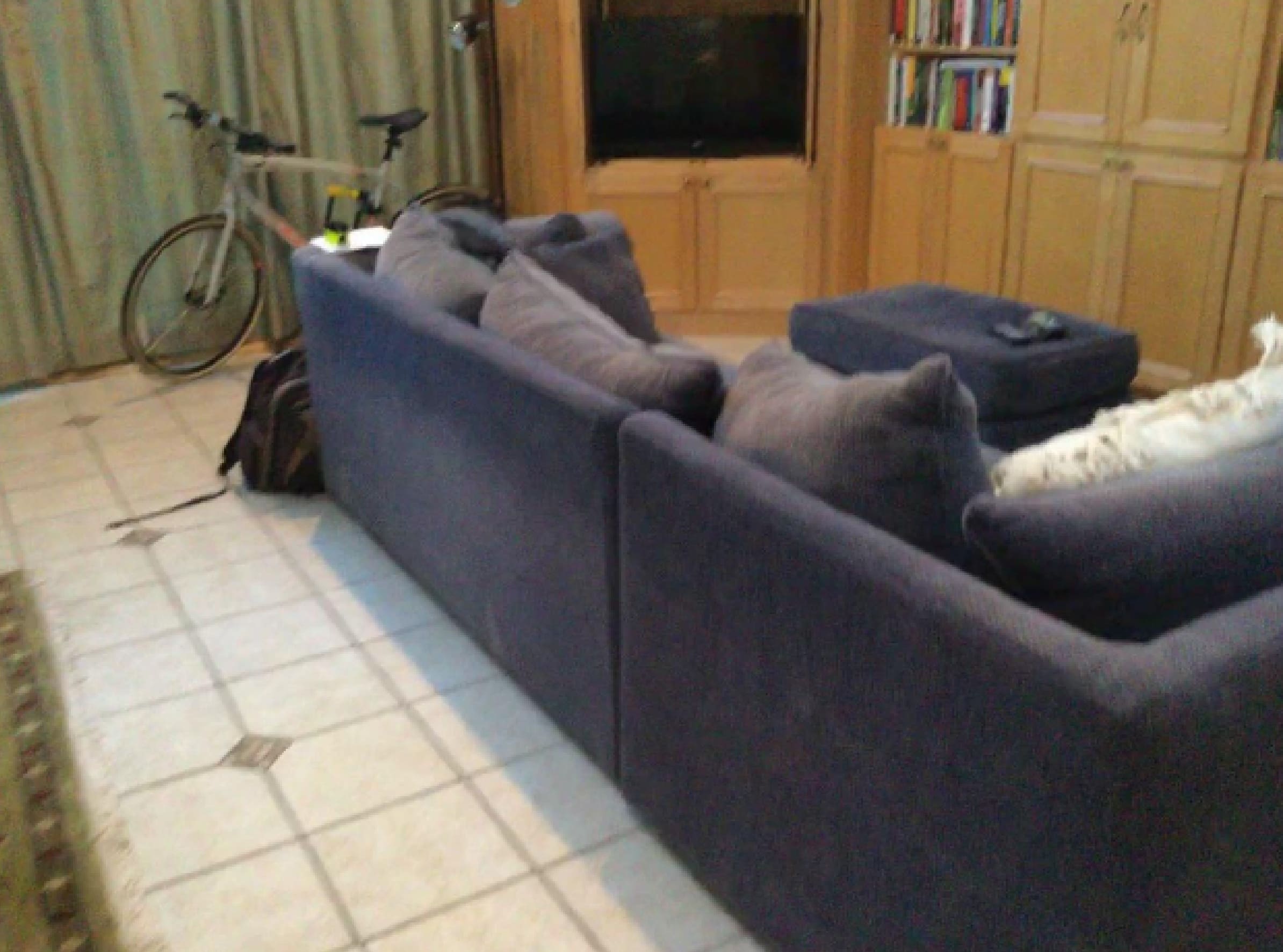}
        \end{minipage}
    } 
       
\centering
\caption{\textbf{Rendering results on the ScanNet dataset.} 
}
\label{scannet color}
\end{figure}

\subsubsection{Evaluation on TUM RGB-D \cite{sturm12iros}}
We evaluated our method's camera tracking performance on the TUM RGB-D dataset, which is of a small scale. Our method significantly outperformed implicit neural scene representation methods as demonstrated in TABLE \ref{tum_tracking}. The result shows that the traditional tracking module with loop closure in our system is more accurate than the neural implicit tracking.

\subsubsection{Evaluation on ScanNet \cite{dai2017scannet}}
We evaluated our method's scene representation capabilities on a larger real-world scene dataset ScanNet \cite{dai2017scannet}, which is more challenging than synthetic ones in tracking and mapping. We selected the same scenes as NICE-SLAM.

The quantitative experiment of tracking on ScanNet is shown in TABLE \ref{scannet_tracking}. Our method outperforms baseline methods in most scenes and on average.
The qualitative experiment results on ScanNet are shown in Fig. \ref{scannet color} and the extracted mesh is shown in Fig. \ref{normal mesh}. The images rendered by our method are sharper and more detailed than those produced by NICE-SLAM and ESLAM.

\begin{table}[t]
\centering
\caption{\textbf{Runtime \& Memory usage.} Mapping time refers to the duration of a single iteration.}
\begin{tabular}{lccc}
\toprule
        \makebox[0.08\textwidth]&  \makebox[0.1\textwidth]{Mapping time [ms]} & \makebox[0.08\textwidth]{\texttt{\# Frames}} &\makebox[0.08\textwidth]{Mem. [MB]}\\
\midrule
iMAP \cite{sucar2021imap}  & 448 & 400 & \textbf{1.04}\\
NICE-SLAM \cite{zhu2022nice}  & 130 & 400 & 12.02  \\
ESLAM \cite{johari2022eslam}  & 19 & 400 & 6.79  \\
\textbf{Ours}  & \textbf{14}  & \textbf{120} & 5.32 \\
\bottomrule
\end{tabular}
\label{time table}
\end{table}

\subsection{Performance Analysis}

In the previous subsection, We evaluate the geometry and appearance of scene reconstruction quality and camera tracking accuracy.
However, a good SLAM system does not only perform well in these aspects.
In the following, we evaluate other properties of our method and compare them with other implicit neural SLAM systems.

\subsubsection{Runtime and Memory Usage}
We evaluate the mapping time, the number of mapping frames, and the memory usage on the eight scenes of the Replica dataset in TABLE \ref{time table}. Here, mapping time refers to the duration to process $M=1000$ pixels in a single iteration, which is the same as NICE-SLAM. As shown in the first two columns of TABLE \ref{time table}, our mapping time of each single iteration is the shortest compared to the baselines. Further, our method requires significantly fewer frames for mapping while achieving better mapping performance, as we use only keyframes for mapping. This further speeds up training as the data I/O time is reduced. This allows our system to meet the low-latency requirement, i.e., the majority of information of a keyframe is integrated into the map before receiving a new keyframe. 

Regarding memory usage, iMAP requires the lowest memory footprint as it only uses a single MLP to represent the scenes. Note that our method outperforms all the other baselines thanks to the octree structure.

\begin{figure}[t]
    \subfloat{
         \centering
         \includegraphics[width=0.9\linewidth,trim=4 4 4 4,clip]{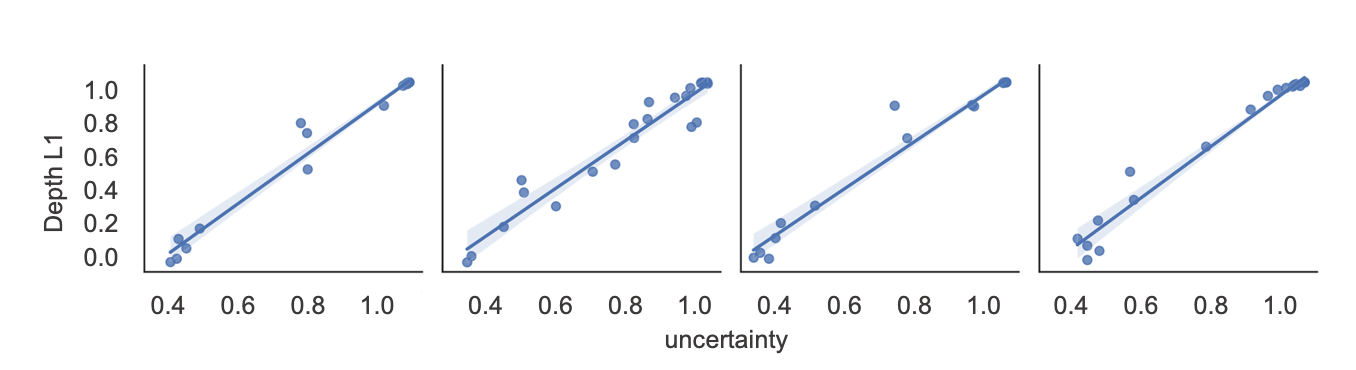}
        }
    
        \subfloat{
         \centering
         \includegraphics[width=0.9\linewidth,trim=4 4 4 4,clip]{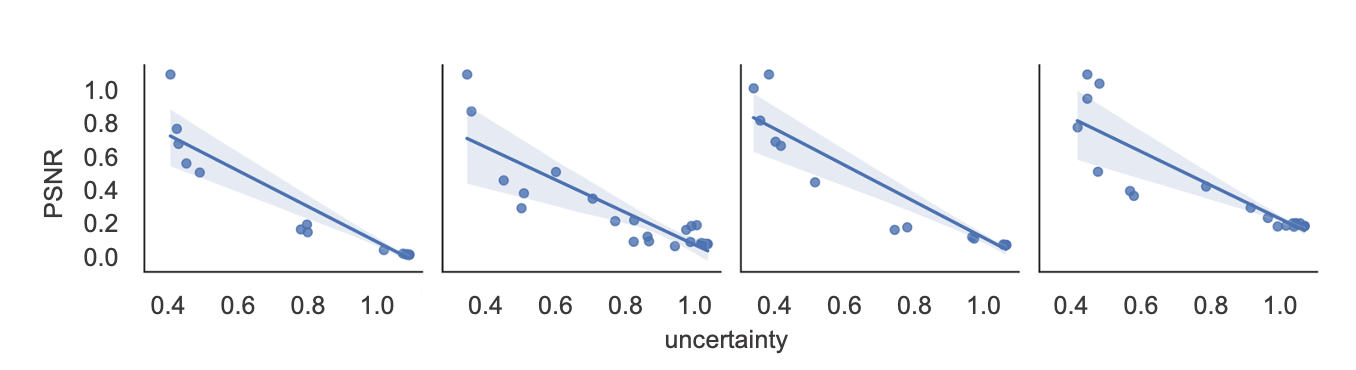}
        }
        
    \caption{\textbf{Relation between uncertainty and depth L1, PSNR} on four images. We normalize depth L1 and PSNR to $[0,1]$ based on their worst and best values for better illustration.}
    \label{rel}
\end{figure}

\begin{figure}[]
\centering   
\captionsetup[subfloat]{labelfont=scriptsize,textfont=scriptsize}
    \subfloat[Before LC]{
    \centering
    
        \begin{minipage}{0.23\linewidth}
            \centering
            \includegraphics[width=\textwidth]{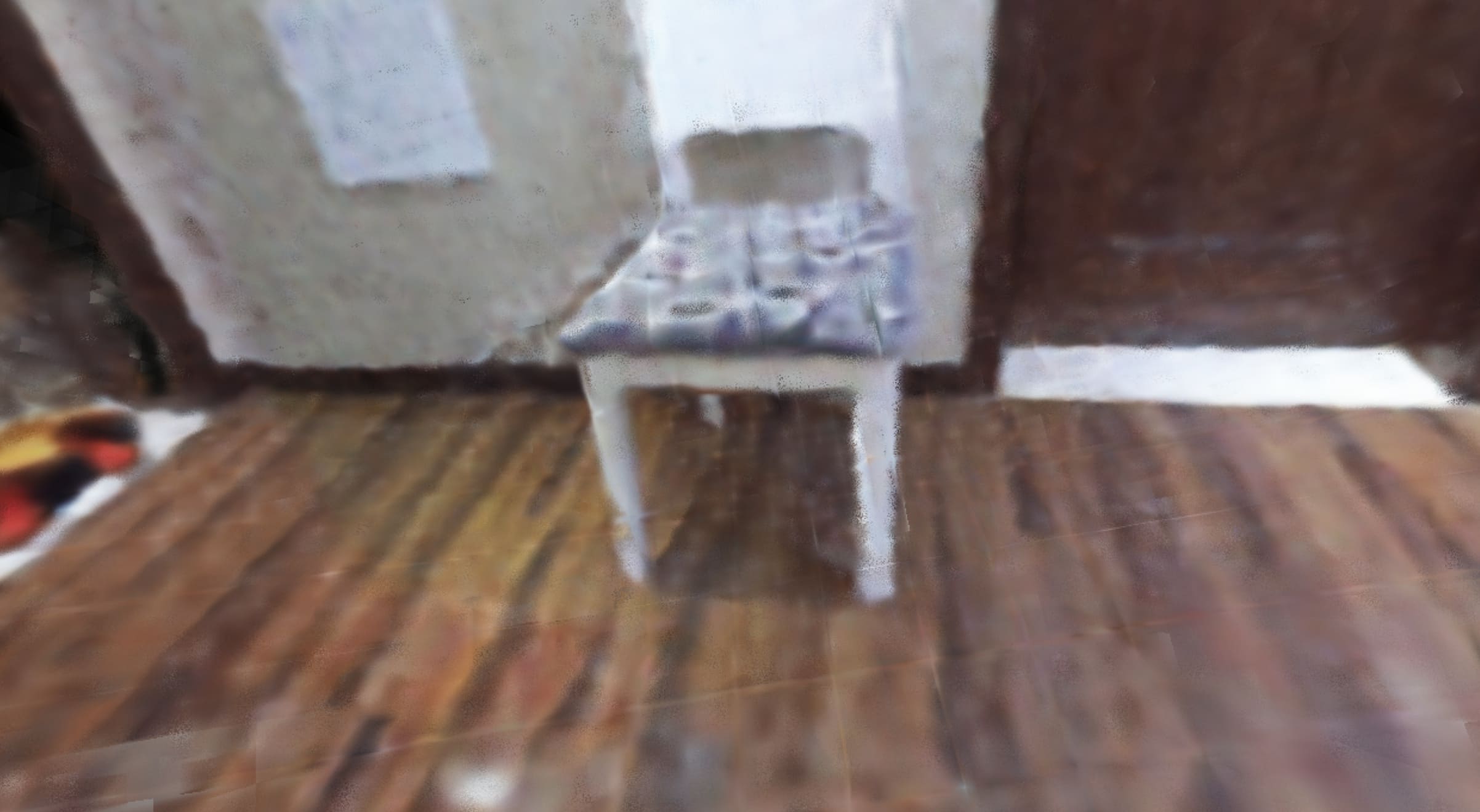}
        \end{minipage}
    }\hspace{-3mm}    
    \subfloat[After LC]{
    \centering
        \begin{minipage}{0.23\linewidth}
            \centering
            \includegraphics[width=\textwidth]{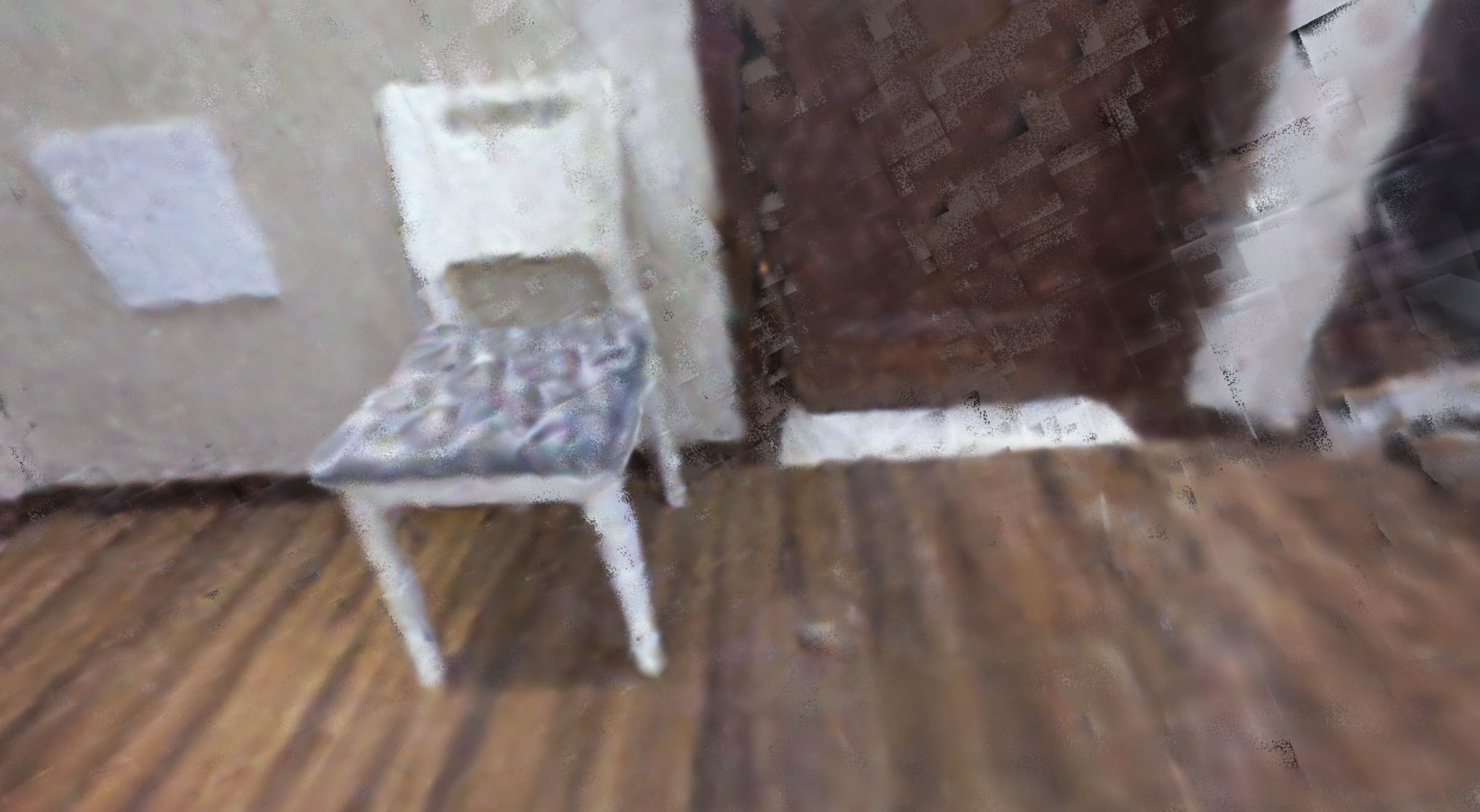}
        \end{minipage}
    }\hspace{-3mm}
    \subfloat[After FT]{
    \centering
        \begin{minipage}{0.23\linewidth}
            \centering
            \includegraphics[width=\textwidth]{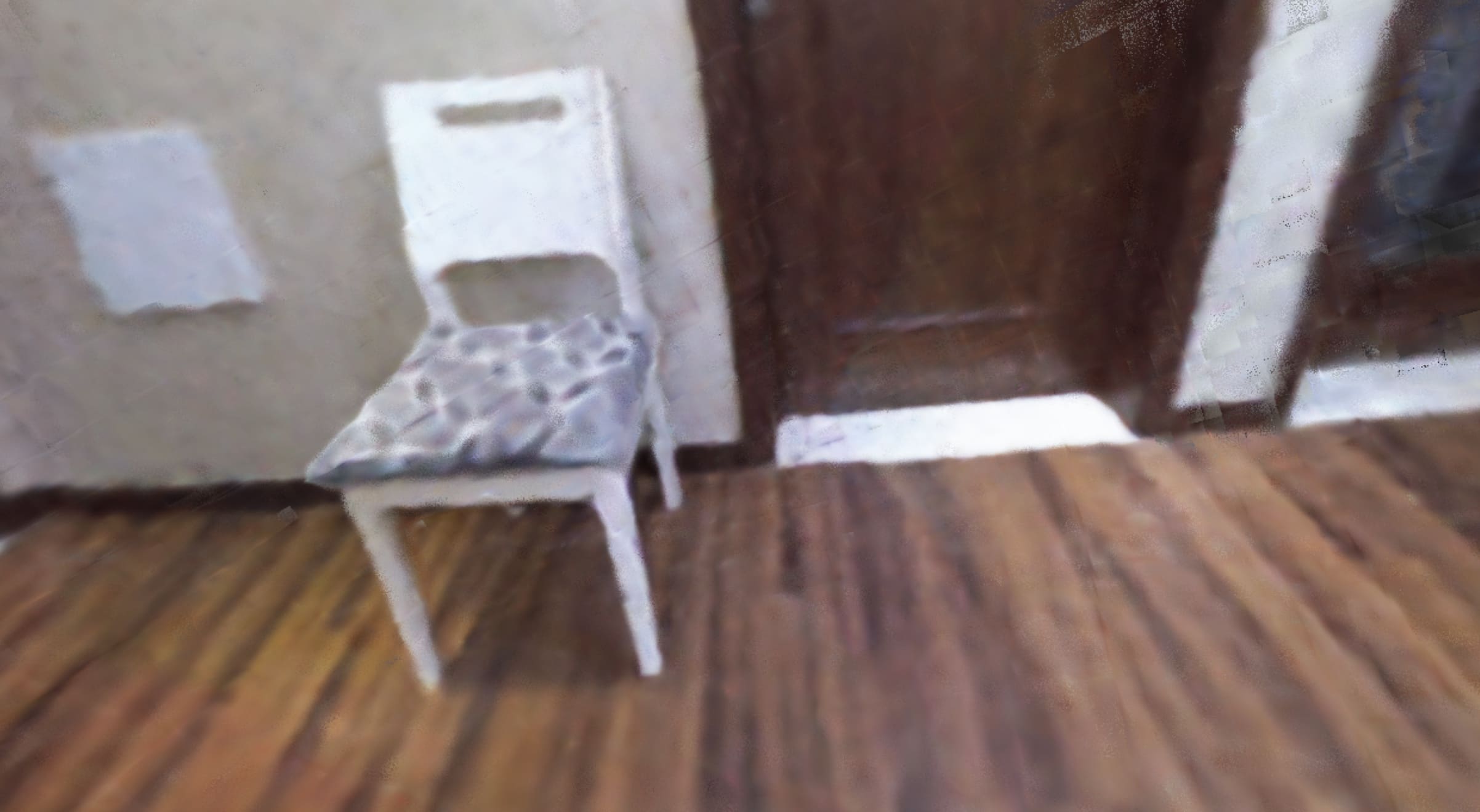}
        \end{minipage}
    }\hspace{-3mm}
    \subfloat[GT]{
    \centering
        \begin{minipage}{0.23\linewidth}
            \centering
            \includegraphics[width=\textwidth]{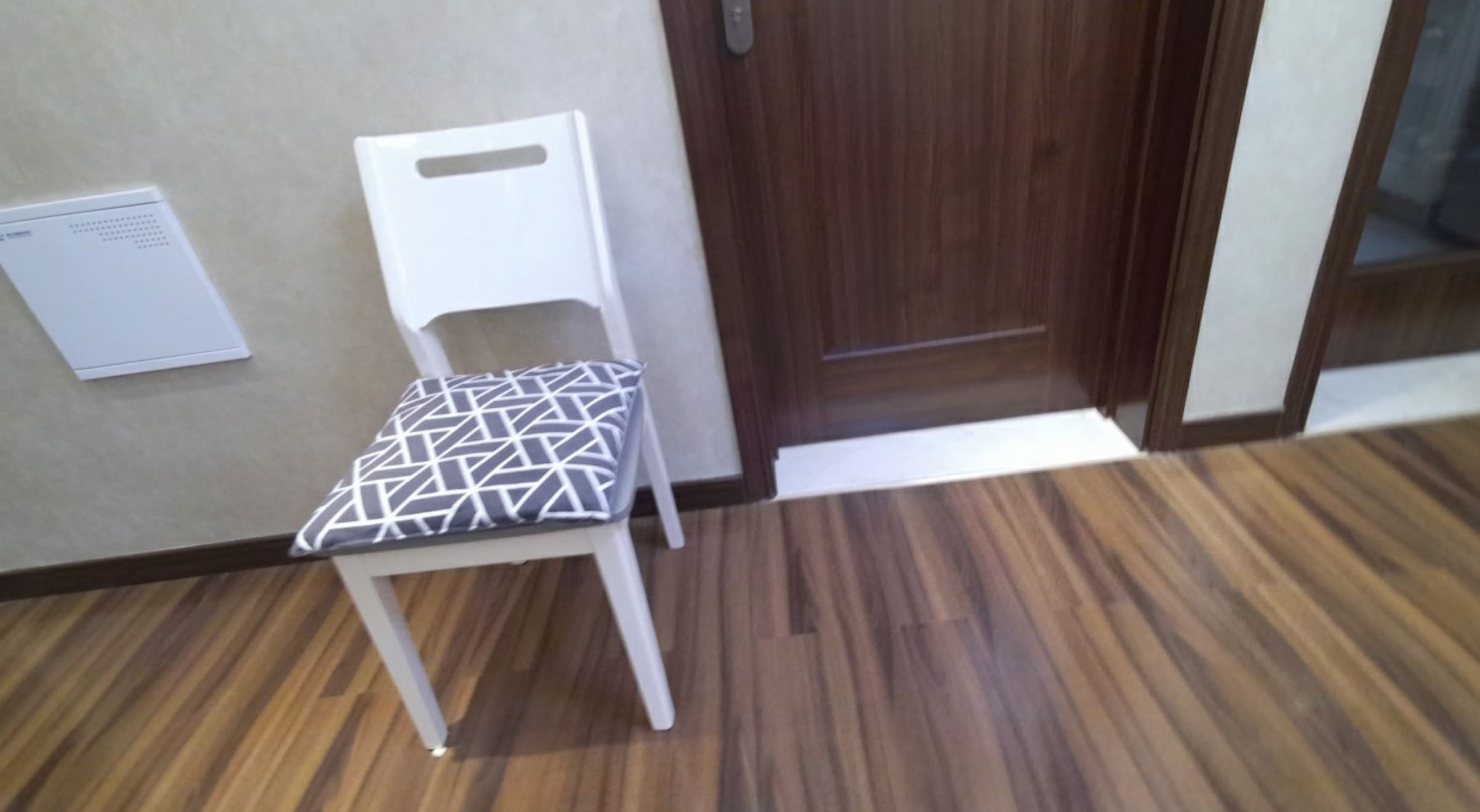}
        \end{minipage}
    }

    \caption{\textbf{Ablation study on the sub-map adjustment and map fine-tuning.} LC refers to loop closure and FT refers to map fine-tuning. 
    }
    \label{local map ablation}
\end{figure}

\subsubsection{Uncertainty}
We propose an uncertainty calculation method in section \ref{render}. The uncertainty is designed to identify pixels where occupancies are not well-trained. 
In this experiment, we investigate the relationship between uncertainty and depth L1 and PSNR. We randomly selected four viewpoints and rendered the depth map and color map of the corresponding viewpoints in each volume. As shown in Fig. \ref{rel}, depth L1 is positively correlated with uncertainty, and PSNR is negatively correlated with uncertainty, indicating our uncertainty-based image selection strategy is effective.

\begin{table}[t]
\centering
\caption{\textbf{Ablation study on the map fine-tuning (FT).} 
}
\begin{tabular}{lcc}
\toprule
        &{Ours w/o FT} & \makebox[0.08\textwidth][c]{Ours} \\
\midrule
Depth L1 [cm] $\downarrow$ &  1.34 & \textbf{1.28}\\
\midrule
PSNR $\uparrow$  &28.39 & \textbf{29.53}\\
SSIM $\uparrow$ & 0.837 & \textbf{0.864} \\
LPIPS $\downarrow$  &0.203 & \textbf{0.156}\\
\bottomrule
\end{tabular}
\label{ablation_gba}
\end{table}

\begin{figure}[t]
    \centering
    \includegraphics[width=0.7\linewidth]{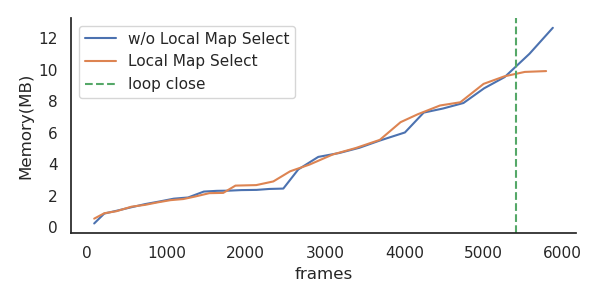}
    \vspace{-0.3cm}
    \caption{\textbf{Ablation study on local map selection.} 
    }
    \label{memory usage ablation}
\end{figure}

\subsection{Ablation study}
In this section, we show the importance of our sub-map adjustment for loop-closing scenes and how useful map fine-tuning and local map selection are.

\subsubsection{Sub-map adjustment}
We verify the effectiveness of sub-map adjustment. 
Given a sequence of trajectory with loop closure, we compared the RGB-D image rendered before and after loop closure. 
As shown in Fig. \ref{local map ablation} (a) and (b), our method with sub-map adjustment can quickly respond to a loop closure without re-training the models and eliminate most of the errors.

\subsubsection{Map Finetuning}
We verify the effectiveness of our map finetuning. We compare our method with and without the map finetuning on the Replica dataset. As shown in TABLE \ref{ablation_gba}, we quantitatively show that the map finetuning improves the performance of our mapping module. 
Also, the rendering images with and without map finetuning are shown in Fig. \ref{local map ablation}. After map finetuning, the quality of rendering images is higher.

\subsubsection{Local Map Selection}
When a new keyframe comes in, our method determines whether it belongs to previous local maps or a new local map is required.
This method ensures no redundant local maps for the same place. As shown in Fig. \ref{memory usage ablation}, the method keeps the memory usage from growing after loop closure.

\section{Conclusions}

In this paper, we propose a global consistent neural implicit representation-based SLAM system,
NGEL-SLAM, for indoor scenes. Combining the traditional tracking and neural implicit scene representation, our method generates high-precision mesh while tracking accurate camera poses. Compared to other neural implicit SLAM systems, our approach ensures global consistency and low latency which is more suitable for real-world applications.

\bibliographystyle{IEEEtran}
\bibliography{IEEEabrv,refs}

\end{document}